\documentclass[preprint,12pt,numbers]{elsarticle} 
\usepackage{kotex}  
\usepackage{multirow}
\usepackage{graphicx}
\usepackage{array} 
\usepackage{tabularx} 
\usepackage{placeins} 
\usepackage{geometry} 
\usepackage{adjustbox}
\usepackage{hyperref} 
\usepackage{listings}
\usepackage{makecell}
\usepackage{float} 
\usepackage[dvipsnames]{xcolor}  
\usepackage{pifont}   
\usepackage{enumitem}
\usepackage{booktabs} 
\usepackage{caption}
\usepackage[table,dvipsnames]{xcolor}
\usepackage{tcolorbox}
\usepackage{amssymb}
\usepackage{amsmath}

\newcommand{\cmark}{{\color{LimeGreen}\ding{51}}}
\newcommand{\xmark}{{\color{red}\ding{55}}}           
\newcommand{\muscad}{\textsc{MUSCAD}}

\definecolor{cBlue_1}{RGB}{115,186,214}
\definecolor{cBlue_6}{RGB}{13,76,109}
\definecolor{cBlue_7}{RGB}{16,106,130}
\definecolor{cBlue_8}{RGB}{19,136,160}
\definecolor{green}{HTML}{E0E5B6}

\journal{Knowledge-Based Systems}

\begin{document}

\begin{frontmatter}

\title{A Scalable Unsupervised Framework for multi-aspect labeling of Multilingual and Multi-Domain Review Data}

\author[a]{Jiin Park}
\ead{jiinpark@hanyang.ac.kr}

\author[a,b]{Misuk Kim\corref{cor1}} 
\cortext[cor1]{Corresponding author}
\ead{misukkim@hanyang.ac.kr} 

\affiliation[a]{
  organization={Department of Artificial Intelligence, Hanyang University},
  addressline={222 Wangsimni-ro, Seongdong-gu},
  city={Seoul},
  postcode={04763},
  country={Republic of Korea}
}
\affiliation[b]{
  organization={Department of Data Science, Hanyang University},
  addressline={222 Wangsimni-ro, Seongdong-gu},
  city={Seoul},
  postcode={04763},
  country={Republic of Korea}
}
\begin{abstract}

Effectively analyzing online review data is essential across industries. However, many existing studies are limited to specific domains and languages or depend on supervised learning approaches that require large-scale labeled datasets. To address these limitations, we propose a multilingual, scalable, and unsupervised framework for cross-domain aspect detection. This framework is designed for multi-aspect labeling of multilingual and multi-domain review data. 
In this study, we apply automatic labeling to Korean and English review datasets spanning various domains and assess the quality of the generated labels through extensive experiments. Aspect category candidates are first extracted through clustering, and each review is then represented as an aspect-aware embedding vector using negative sampling. To evaluate the framework, we conduct multi-aspect labeling and fine-tune several pretrained language models to measure the effectiveness of the automatically generated labels. Results show that these models achieve high performance, demonstrating that the labels are suitable for training. Furthermore, comparisons with publicly available large language models highlight the framework’s superior consistency and scalability when processing large-scale data. A human evaluation also confirms that the quality of the automatic labels is comparable to those created manually. 
This study demonstrates the potential of a robust multi-aspect labeling approach that overcomes limitations of supervised methods and is adaptable to multilingual, multi-domain environments. Future research will explore automatic review summarization and the integration of artificial intelligence agents to further improve the efficiency and depth of review analysis.

\end{abstract}

\begin{graphicalabstract}
\includegraphics[width=\textwidth,keepaspectratio]{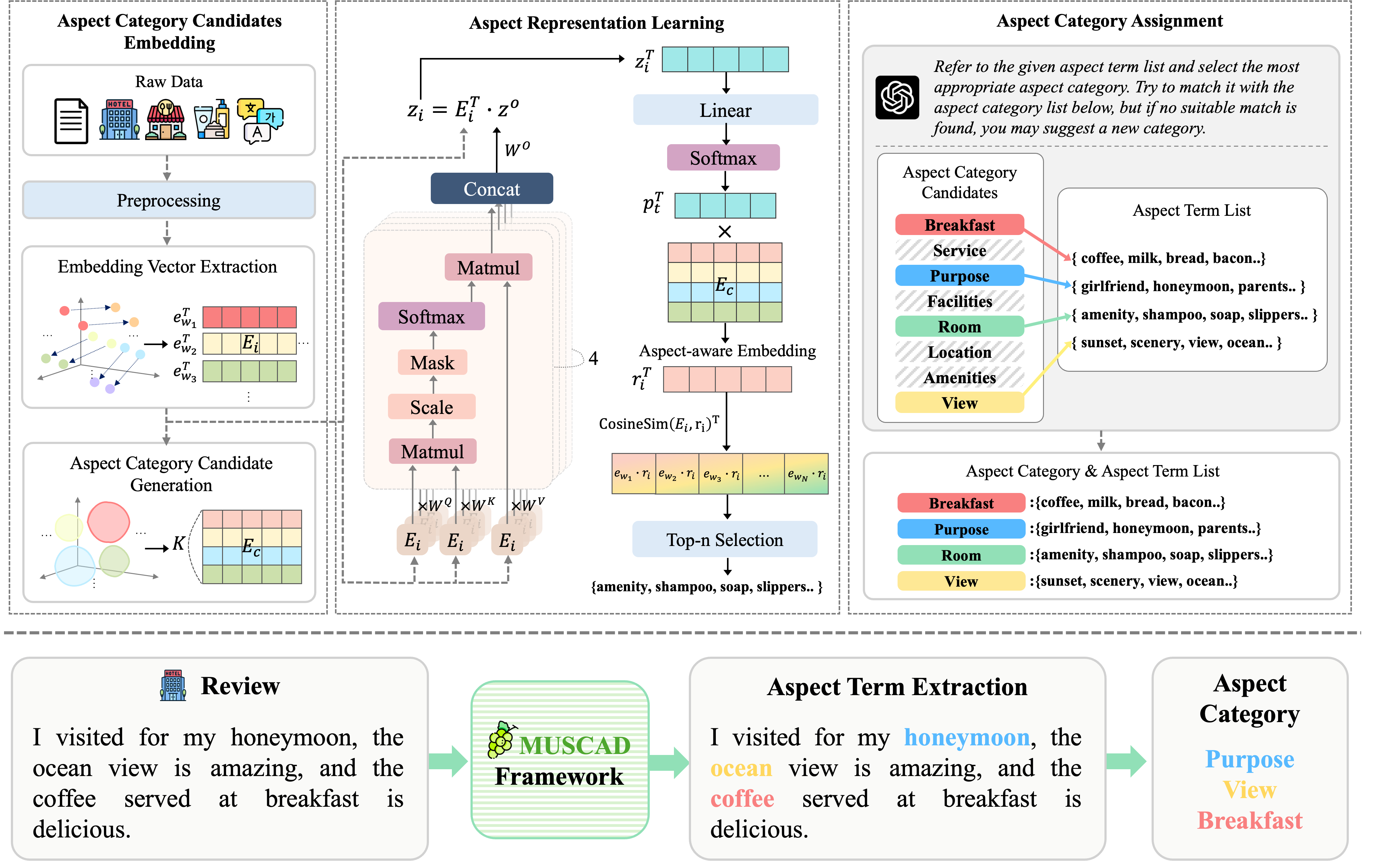}
\end{graphicalabstract}

\begin{highlights}
\item Proposes a scalable framework for multi-aspect category labeling without supervision.
\item Discovers semantically similar aspect terms and clusters them without predefined labels.
\item Learns aspect semantics using multi-head attention, negative sampling, and clustering.
\item Assigns aspect categories using prompts in zero- and few-shot settings with language models.
\item Achieves strong performance across domains and languages, including English and Korean.
\end{highlights}


\end{frontmatter}

\section{Introduction}

User review data contains direct feedback from consumers about products and services, and can be used as an important source of information for business decision-making, marketing strategy development, and brand reputation management. However, most review data exists in unstructured text format, and typical sentiment polarity analysis or star rating–based analysis has the limitation of not sufficiently capturing the detailed attributes embedded in the reviews. As an alternative to overcome these limitations, Aspect-based Analysis has been gaining attention. Aspect-based Analysis can be widely used in various fields such as consumer feedback analysis, public policy monitoring, and social media analysis~\citep{Khabiri_Caverlee_Hsu_2021,qiu-etal-2011-opinion}, and has the advantage of enabling detailed understanding of how each aspect is evaluated in text.

However, even Aspect-based Analysis has several limitations when applied to user review data, which can be broadly categorized into four aspects.  
First, most studies on Aspect-based Analysis focus on specific domains such as electronics or restaurant reviews~\citep{zhao-etal-2020-spanmlt}, making it difficult to ensure that the proposed methods show the same performance and validity in real service environments where a wider range of domains must be addressed. In other words, studies conducted within limited domains do not sufficiently demonstrate multi-domain applicability, which restricts their practical use across industries.  
Second, Aspect-based Analysis research tends to focus on quantitative metrics such as accuracy and F1-score~\citep{zhang-etal-2024-self-training}, and lacks qualitative research verifying whether the aspects extracted by the model are semantically valid and practically effective in the real world. As a result, it becomes difficult to determine whether the extracted aspects are trustworthy enough to be used for problem diagnosis or product improvement in practice, which again acts as a limitation in applying the methods in industry.  
Third, most Aspect-based Analysis research adopts approaches that fine-tune labeled data or assume specific data formats, which leads to high labeling costs and necessitates format conversion~\citep{mukherjee-etal-2021-paste}. When relying solely on labeled data, it becomes difficult to accurately capture aspects from newly appearing review texts that are not labeled, leading to limitations in responding in real time in actual service environments.  
Lastly, another issue is that most Aspect-based Analysis studies are conducted mainly on English data. Since large-scale review data is accumulated in various languages in global service environments, models trained only in a specific language face limitations in expanding to multilingual data. As a result, reviews written in newly emerging or underrepresented languages may not be sufficiently analyzed, leading to missed opportunities for satisfaction analysis and service improvement in various markets.

Therefore, this study aims to overcome the limitations of existing Aspect-based Analysis and proposes \muscad~(\textbf{Mu}ltilingual and \textbf{S}calable framework for \textbf{C}ross-domain \textbf{A}spect \textbf{D}etection), an unsupervised multi-aspect labeling framework for multilingual and multi-domain review data. The core ideas of \muscad~are as follows:
\begin{enumerate}[label=(\alph*)]
\item Without large-scale labeling costs or domain-specific formatting, K-means clustering is applied directly to unlabeled review texts to automatically group semantically similar words and generate initial aspect category candidates.
\item To extract aspect-aware embedding vectors that reflect contextual dependencies even across diverse sentence structures, Multi-Head Attention is applied to finely learn key information within sentences, and \texttt{Max-Margin Loss} is used to sharpen the boundaries between aspects.
\item Finally, to ensure scalability across multiple domains and languages, the aspect category candidates extracted through K-means are refined by leveraging both domain expert knowledge and \texttt{GPT}-based models to perform-domain and language-specific naming with improved readability.
\end{enumerate}

To evaluate the effectiveness of \textsc{MUSCAD}, we first perform large-scale automatic labeling on hotel, food, and beauty review datasets written in Korean and English. Next, to verify whether the labels generated by the unsupervised method are practically usable in downstream tasks or supervised learning, we fine-tune classification models and compare performance in terms of accuracy (F1-score). Additionally, we prompt large language models to perform multi-aspect category labeling on the same dataset, and compare the results with the automatically generated labels to verify how consistent and accurate they are compared to LLMs. Lastly, to qualitatively confirm the semantic validity of the extracted labels, we conduct Human Evaluation and compare the results with expert-annotated labels. Through this series of experiments, we aim to demonstrate that the proposed framework can effectively operate in multilingual and multi-domain environments and generate automatically labeled data of sufficient quality for real-world use.

The main contributions of this paper are summarized as follows:

\begin{itemize}
\item \textbf{Proposal of an unsupervised multi-aspect labeling framework:} We propose a new framework that automatically extracts and groups multi-aspect category without manual labeling.
\item \textbf{Verification of multilingual and multi-domain scalability:} We conduct experiments on Korean and English review datasets and empirically validate the framework’s applicability to diverse domains.
\item \textbf{Quantitative and qualitative performance evaluation:} We assess the reliability of the automatically labeled data through comparison with pre-trained classifiers and conduct qualitative evaluation through human annotation.
\item \textbf{Provision of multi-aspect labeled datasets:} We construct high-quality automatically labeled datasets and provide them for use in various downstream tasks through fine-tuning.
\end{itemize}


\section{Related Work}

This section reviews previous research on Aspect-based Analysis and examines studies related to its applicability across multiple languages and domains. It also covers recent trends in attention mechanisms and their applications.

\subsection{Aspect-based Analysis}

Research focusing on the analysis of specific aspects in various types of text data has been actively conducted. This technique is utilized in diverse fields such as consumer review analysis, public opinion monitoring, stock market trend prediction~\citep{Lee:JIIS:2018:749,Lee:JIIS:2018:756, Park:JIIS:2019:798}, and online marketing strategy development~\citep{10.1145/3397271.3401269,angelidis-etal-2021-extractive}.

Early studies mainly relied on manually defined rule-based approaches, applying topic modeling methods such as Latent Dirichlet Allocation (LDA) to extract aspect terms from text~\cite{Khabiri_Caverlee_Hsu_2021}. \cite{qiu-etal-2011-opinion} proposed extracting aspects using part-of-speech tagging and word frequency data. Another method involved setting preprocessing rules, performing frequency analysis on text data, and manually assigning keywords to predefined categories based on domain knowledge to evaluate keyword importance per attribute~\citep{ART003131848}. \cite{zhao-etal-2020-spanmlt} proposed a method to extract aspect terms and opinion terms simultaneously, but since it mainly focused on identifying individual elements, it had limitations in effectively extracting aspect-opinion pairs. These approaches depend on predefined rules, making them difficult to apply to various domains, and because aspect categories are selected based on domain knowledge, they have limitations in reflecting domain-specific terms. Moreover, since most analyses are limited to nouns, it is difficult to conduct sentiment analysis that reflects context.

To overcome these limitations, supervised learning–based approaches have been introduced. Early studies explored the use of Conditional Random Fields (CRFs) to simultaneously extract structural features of sentences and polarity opinions~\citep{li-etal-2010-structure}. More recently, deep neural network–based approaches have shown promising performance. \cite{wang-etal-2016-attention} applied an attention-based Long Short-Term Memory (LSTM) model for aspect-level sentiment classification, allowing the model to focus on important words in the sentence based on the input aspect. \cite{fan-etal-2019-target} defined a method called Target-oriented Opinion Words Extraction (TOWE) using Inward-Outward LSTM to extract opinion words related to specific targets. Additionally, the RNCRF model, which combines Recurrent Neural Networks (RNNs) and CRFs, was introduced to learn the relationship between aspects and opinions, and a joint learning method was proposed by optimizing CRF parameters~\citep{wang-etal-2016-recursive}. Topic-level sentiment analysis research has also been conducted. Approaches using Online Latent Semantic Indexing and LSTM-based models have been proposed to classify sentiment for specific topics in sentences~\citep{PATHAK2021107440}. \cite{mikolov-etal-2013-linguistic} conducted research on learning word representations using neural network–based language models. However, supervised deep learning models require a large amount of labeled data, which involves high cost and time for data construction. Moreover, models trained in a specific domain are difficult to apply directly to other applications.

To address these limitations, recent studies have drawn attention to unsupervised learning–based approaches, which aim to perform effective aspect extraction and sentiment analysis without labeled data. Local topic models at the sentence level have been proposed to automatically infer aspects and generate seed sets of sentiment words without manual annotation~\citep{10.5555/1857999.1858121}. \cite{mukherjee-liu-2012-aspect} proposed a semi-supervised topic model that jointly models aspects and sentiment. \cite{zhao-etal-2010-jointly} combined topic modeling and maximum entropy (MaxEnt) models to separate aspect and opinion words using a semi-supervised approach. More recently, unsupervised aspect term extraction using attention-based Deep Neural Networks (DNNs) has been proposed~\citep{he-etal-2017-unsupervised}. These approaches have the advantage of automatically clustering semantically similar aspects without labeled data. However, existing unsupervised methods lack sufficient verification of cross-domain generalization performance, and due to instability during training, they often suffer from low reproducibility and inconsistent performance.

Most previous studies have been conducted in single-language settings. For example, in Aspect-Based Sentiment Analysis using Korean data, \cite{ART002668954} explored Attribute Category Sentiment Classification (ACSC) using Bidirectional Encoder Representations from Transformers (BERT). They replaced the [CLS] token with output vectors from tokens related to attribute categories and analyzed performance differences according to input configurations (QA/NLI). In a study using beauty commerce reviews, Importance-Performance Analysis (IPA) and DistilBERT-based sentiment analysis were used to identify areas for product improvement~\citep{ART003131848}. Another study analyzed sentiment for specific evaluation categories such as food, price, service, and atmosphere using Korean restaurant reviews and proposed a model for predicting detailed ratings~\citep{ART002596643}. \cite{ART002826299} applied attribute-based sentiment analysis using BERT in a movie recommendation system to classify reviews by attributes (e.g., direction, actors, story) and develop a Movie Multi-Criteria recommendation model reflecting user preferences. These studies demonstrate that Aspect-Based Sentiment Analysis enables more fine-grained evaluation of products and services and can enhance the performance of recommendation systems. However, they still lack sufficient verification of generalization across domains and studies considering multilingual applicability remain scarce.

\subsection{Attention Mechanism in Aspect-based Analysis}

Recently, attention mechanisms have been widely used in Natural Language Processing (NLP), significantly enhancing the performance of deep neural network models. In particular, pre-trained models such as BERT, Robustly Optimized BERT Pretraining Approach (RoBERTa) effectively capture contextual information through attention, achieving state-of-the-art performance in various NLP tasks. 
Attention mechanisms learn by focusing on important information within the text. For example, in sequence tagging tasks, they highlight specific words within context~\citep{10.1145/3459637.3482186}, and in sequence-to-sequence learning, they help maintain core information during sentence transformation~\citep{10.1609/aaai.v33i01.33018778}. In sentiment analysis, attention helps models perform more precise predictions by focusing on words related to specific aspects~\citep{WANG2021113603}, and in machine translation, it contributes to improving semantic alignment between source and target languages~\citep{zhang-etal-2023-empirical}. Many researchers have explored the use of attention mechanisms in supervised models for effective Aspect-Based Sentiment Analysis (ABSA)~\citep{xu-etal-2019-bert}. Additionally, unsupervised attention-based models have been proposed to extract aspects from text without large-scale labeled data~\citep{he-etal-2017-unsupervised}. However, most of these existing studies focus on single languages and specific domains, and there is still a lack of research verifying generalization performance in multilingual and multi-domain environments.


\section{Dataset}

In this study, we collected review data written in Korean and English to evaluate the performance of our framework in multilingual and multi-domain settings. Specifically, we gathered hotel and food reviews in Korean and beauty reviews in English, applying different preprocessing pipelines according to the linguistic characteristics of each language.

For Korean review data, we used the Korean String Processing Suite (KSS)\footnote{\url{https://github.com/hyunwoongko/kss}} to segment the text into individual sentences for sentence-level analysis. As a result, we obtained 313,333 sentence-level entries for hotel reviews and 288,874 for food reviews. After removing unnecessary special characters and extra spaces, we refined the text using the \texttt{Open Korean Text (OKT)} morphological analyzer from Konlpy~\citep{park2014konlpy}.
Since Korean features various grammatical transformations including verb and adjective inflections, we lemmatized these parts of speech and removed case particles (e.g., Korean case particles such as \textit{i/ga} (subject), \textit{eul/reul} (object), \textit{e} (to/at), \textit{eseo} (from/in)) to ensure analytical consistency. For example, the Korean expressions \textit{\textquotedblleft 맛있었다\textquotedblright} (delicious + past tense) and \textit{\textquotedblleft 맛있을 것이다\textquotedblright} (delicious + future tense) were both normalized to \textit{\textquotedblleft 맛있다\textquotedblright} (delicious). This preprocessing step was critical for reducing variability caused by Korean-specific tense and ending changes that could interfere with model training. We also customized the OKT user dictionary by adding neologisms, slang, and domain-specific proper nouns. This helped prevent incorrect token splits of words not included in the default dictionary and improved the performance of the analyzer. Both preprocessed and original sentences were retained to enable flexible follow-up analyses without data loss.

For English review data, we used the \texttt{sent\_tokenize} function from the NLTK library\footnote{\url{https://github.com/nltk/nltk_data}} to segment the text into sentences. All text was lowercased, and special characters were removed. Then, stopwords were filtered out to eliminate words irrelevant to the analysis. Since English features relatively less morphological variation than Korean, we only applied stopword removal and word normalization without additional lemmatization steps.
As a result of these preprocessing steps, we constructed a dataset covering three domains: hotel, food, and beauty, as shown in Table~\ref{tbl1}.

\begin{table}[t]
\centering
\footnotesize
\begin{tabular}{@{}l l r r@{}}
\noalign{\hrule height 1.3pt}
\toprule
\textbf{Domain} & \textbf{Language} & \textbf{Number of Reviews} & \textbf{Number of Sentences} \\
\midrule
Beauty & English & 50,000 & 255,110 \\
Hotel  & Korean  & 225,412 & 313,333 \\
Food   & Korean  & 192,484 & 288,874 \\
\bottomrule
\noalign{\hrule height 1.3pt}
\end{tabular}
\caption{Overview of the Hotel, Food, and Beauty dataset with review and sentence counts for each domain in English and Korean.}
\label{tbl1}
\end{table}

\section{Framework Structure}
\label{sec:framework_structure}

\begin{figure}[htbp]
    \centering
    \includegraphics[width=1.07\linewidth]{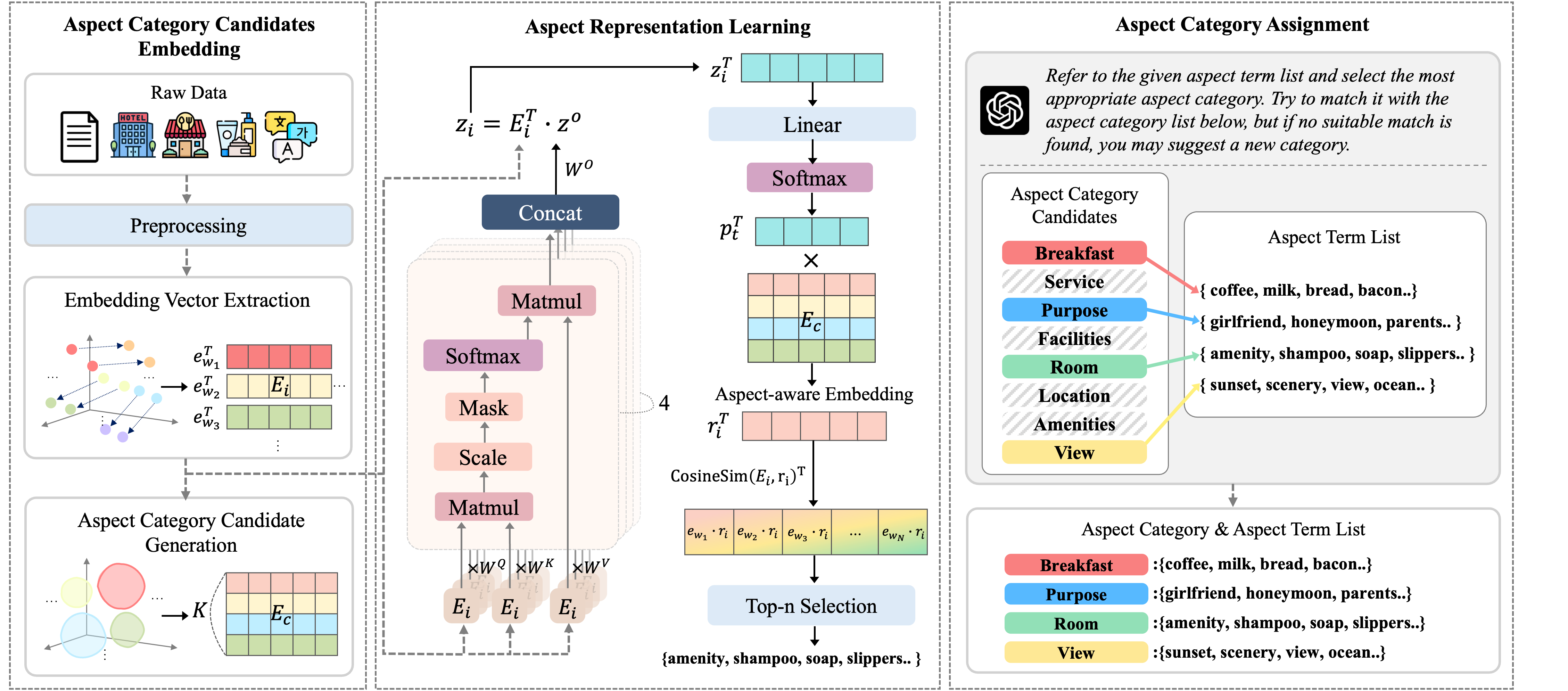} 
    \caption{\muscad~ Framework}
    \label{FIG:1}
\end{figure}

\muscad~is designed to automatically extract aspect terms and aspect categories from sentences and to learn their interrelationships. Its architecture incorporates K-means clustering, aspect representation learning through self-attention and multi-head attention, and aspect classification optimization using max-margin loss. The overall structure of the proposed framework is shown in Fig.~\ref{FIG:1}.

Additionally, the extracted aspect terms and categories can be used to automatically assign aspect category labels to new review data, enabling more precise aspect-aware labeling. Section~\ref{sec:framework_structure} describes each sub-module in detail.

\subsection{Aspect Category Candidate Embedding}
\label{sec:4.1}

In the input processing stage, we perform preprocessing to effectively extract aspect terms from natural language sentences. In this study, we apply a method that learns word embeddings without predefined vocabularies and automatically derives initial aspect category candidates through K-means clustering. This enables scalable extraction of aspect terms and aspect categories that are not domain-dependent.

\subsubsection{Embedding Vector Extraction}
\label{sec:4.1.1}

We train word embeddings reflecting semantic similarity using the \texttt{Word2Vec} model \citep{mikolov-etal-2013-linguistic}. Specifically, we apply the Continuous Bag of Words (\texttt{CBOW}) approach, which predicts a center word based on its surrounding context words. For example, in the sentence ``The hotel has great service and staff'', if ``service'' is the center word and the window size is 1, the context words are ``great'' and ``and''.

Each context word is represented as a one-hot encoded vector $x_i$, which is multiplied by a weight matrix to obtain an embedding vector. The average of these context embeddings becomes the context vector, which is passed through the output weights and a \texttt{Softmax} function to predict the probability of the center word. By minimizing the loss function during training, we obtain the embedding matrix $E_i$, which contains embedding vectors for all words.

\subsubsection{Unsupervised Generation of Aspect Category Candidate}
\label{sec:4.1.2}

In the unsupervised aspect extraction stage, since specific aspect terms or categories are not predefined, we must effectively group semantically similar words. We propose applying K-means clustering to the word embeddings generated in Section~\ref{sec:4.1.1} to automatically generate initial aspect category candidates, which are later refined into final aspect categories.

K-means clustering is a representative unsupervised learning technique for grouping data points, and it can automatically cluster word embedding vectors based on their semantic similarity. The result of clustering the data using the K-means algorithm can be represented as shown in Eq.~\ref{eq1}.

\begin{equation}
    C_k = \{ e_{w_i} \ | \ \arg\min_{k} \| e_{w_i} - c_k \|^2 \}, 
    \label{eq1}
\end{equation}
where $e_{i}$ denotes the vector representation of each word, $c_k$ is the centroid of the $k^{th}$ cluster, and $C_k$ represents the set of words belonging to the $k^{th}$ cluster. Each time a word is reassigned to a new cluster, the cluster centroid $c_k$ is updated and then $L_2$-normalized to maintain a consistent scale (see Eq.~\ref{eq2}).

\begin{equation}
    \hat{c}_k = \frac{c_k}{\| c_k \|_2},~~where~~c_k = \frac{1}{|C_k|} \sum_{e_{w_i} \in C_k} e_{w_i}
    \label{eq2}
\end{equation}

Each cluster centroid $c_k$ obtained by the K-means algorithm is computed as the mean of the word embeddings $e_{w_i}$ belonging to cluster $C_k$, and is subsequently $L_2$-normalized into a unit vector $\hat{c_k}$, as shown in Eq.~\ref{eq2}.
In this study, the set of normalized centroids ${\hat{c_1}, \hat{c_2}, \dots, \hat{c_K}}$ is concatenated into a single matrix, which serves as the initial value of the aspect embedding matrix $E_c$.
This design enables the semantic structure captured during clustering to be directly transferred to the aspect representation learning stage

By repeating this process, we group word vectors into semantically coherent clusters and derive optimal initial aspect category candidates based on the centroid vectors. This allows the model to learn relationships between words and aspect categories effectively without predefined vocabularies or labels in multilingual and multi-domain environments.

\subsection{Aspect-Aware Representation Learning}
\label{sec:4.2}

To learn the relationship between each word and aspect category and reflect contextual information in the sentence, we train attention mechanisms. Section~\ref{sec:4.2} describes how we use self-attention and Multi-Head Attention to extract fine-grained, context-aware embedding vectors.

\subsubsection{Multi-Head Attention-Based Sentence Representation}
\label{sec:4.2.1}

Attention mechanisms learn the relationships between words in a sentence and reflect the importance of each word in context. Unlike earlier studies~\citep{he-etal-2017-unsupervised} that relied only on attention, we use Multi-Head Attention to learn sentence representations that capture contextual information.

In this study, the input embedding $E_i$ is fed into the Multi-Head Attention module, where trainable weight matrices are applied to compute the Query, Key, and Value representations, followed by the attention operation.  
The computation process of the Multi-Head Attention is shown in Eq.~\ref{eq3}.
\begin{equation}
    \texttt{MultiHead}(E_i, E_i, E_i) = \texttt{Concat}(\text{head}_1, .., \text{head}_h) W^O,
    \label{eq3}
\end{equation}

$W^O$ is the output weight matrix that projects the concatenated outputs of all attention heads into the final output dimension. Here, $head$ denotes an individual attention head, and each attention head is computed as shown in Eq.~\ref{eq4}.

\begin{equation}
    \text{head}_i= \texttt{softmax} \left( \frac{E_iW_i^Q (E_iW_i^K)^T}{\sqrt{d_k}} \right)E_i W_i^V,  
    \label{eq4}
\end{equation}

Here, $W^Q_i$ projects the input embedding $E_i$ into the query space in the $i^{th}$ attention head, $W^K_i$ projects $E_i$ into the key space within the same head, and $W^V_i$ projects $E_i$ into the value space. Through this Multi-Head Attention process, the final context vector $z^o$ is generated, capturing the relationships between words in the sentence via weighted summation.
While previous studies often derived the sentence representation by directly aggregating the outputs of the Multi-Head Attention, our approach refines contextual representation by multiplying the original word embedding $E_i$ with the context vector $z^o$, resulting in the final sentence representation vector $z_i$.

The resulting $z_i$ represents the core meaning of the sentence and is utilized in the subsequent learning process to extract aspect-aware embeddings.

\subsubsection{Aspect-Aware Embedding Vector}
\label{sec:4.2.2}

In this section, the aspect embedding matrix $E_c$ used to compute the aspect-aware embedding vector is initialized with the $L_2$-normalized centroids obtained from K-means clustering, as described in Section~\ref{sec:4.1.2}.
During training, $E_c$ is further refined through the optimization process described in Section~\ref{sec:4.2.3}, allowing the initial unsupervised clusters to adapt to the learned sentence representations.
In this study, the aspect-aware embedding vector $r_i$ is computed as shown in Eq.~\ref{eq5}, by leveraging the aspect embedding matrix $E_c$ calculated in Section~\ref{sec:4.1} and the probability vector $p_t$, which represents the likelihood of the sentence being associated with each aspect category.
\begin{equation}
    r_i = E_c^Tp_t,~where~ p_t = \texttt{softmax}(W_p^T z_i),
    \label{eq5}
\end{equation}

Here, $W_p$ is the weight matrix that maps the sentence representation $z_i$ into an aspect distribution. Through this process, the $k$ aspect candidates initially generated by K-means are further refined during training, and the final aspect-aware embedding vector is determined by allowing the sentence representation $z_i$ to select the most relevant aspect.

\subsubsection{Optimization of Aspect-Aware Representations}
\label{sec:4.2.3}

As described in Section~\ref{sec:4.2.2}, the aspect-aware embedding vector $r_i$ is computed using the aspect embedding matrix $E_c$ and the probability vector $p_t$.
In this section, we describe how $E_c$ and its related parameters are optimized during training.
Specifically, we adopt the \texttt{Max-Margin Loss} to train the aspect-aware embedding representations. In neural language models, the negative sampling technique is widely used to encourage the model to increase the similarity between a sentence and its corresponding aspect category, while reducing the similarity with negative samples.
During the application of the negative sampling method, the model is guided to learn more generalizable patterns by minimizing the inner product between the sentence vector $z_i$, generated in Section~\ref{sec:4.2.1}, and negative samples, while maximizing the similarity with the most relevant aspect. The negative samples $z_n$ are composed of representation vectors of sentences unrelated to the input sentence, which helps prevent overfitting to a specific aspect during training. 
To achieve this, we define the hinge loss function as shown in Eq.~\ref{eq6}.
\begin{equation}
    \mathcal{L} = \min \sum_{i=1}^{m} \max(0, 1 - \langle z_i, r_i \rangle + \langle z_i^{(n)}, r_i \rangle),
    \label{eq6}
\end{equation}
Here, $\langle~,~ \rangle$ denotes the inner product between two vectors, and $m$ is a hyperparameter representing the number of negative samples selected for each sentence. To enhance the generalization performance of the model, all vectors are $L_2$-normalized to eliminate the influence of vector magnitude.
By training the model based on this loss function, it increases the likelihood of assigning a sentence to the correct aspect while reducing similarity to negative samples. This process minimizes the correlation between different aspects, enabling the model to learn more distinguishable aspect representations.
Ultimately, the model is trained to minimize the \texttt{Max-Margin Loss}, which guides each input sentence to be assigned to the most appropriate aspect category. Through negative sampling, the model can effectively extract aspect categories and aspect terms across diverse sentence structures, enabling more fine-grained, context-aware analysis.

\subsection{Aspect Category Assignment}

In this section, we refine the aspect category candidates derived through unsupervised learning into more interpretable aspect categories. Although K-means clustering groups semantically similar words, the resulting clusters are merely collections of related terms and are not directly usable as aspect categories. Therefore, we perform a post-processing step that analyzes the word lists in each cluster and converts them into semantically meaningful aspect categories.
For example, a cluster in hotel reviews containing words such as ``modern'', ``cozy'' and ``sophisticated'' can be assigned to an aspect category representing room ambiance. Similarly, a cluster in food reviews including terms like ``salty'', ``nutty'' and ``tender'' can be interpreted as referring to the taste of the food. To facilitate this process, we consulted Korean hotel and food review websites as well as international beauty product review sites to construct a set of candidate aspect categories. Then, using \texttt{GPT-3.5-turbo}, we automatically recommended the most appropriate aspect category for each aspect term cluster.
Few-shot prompting was applied based on prompts such as the one shown in Fig.~\ref{fig:aspect_category_prompt}. Fig.~\ref{fig:aspect_category_prompt} provides an example of an aspect categorization prompt for hotel reviews, while aspect category candidates for the food and beauty domains are shown in the Appendix Fig.~\ref{fig:food_beauty_aspect_category_prompt}.

\begin{center}
    \begin{tcolorbox}
        [title={Aspect Categorization Prompt (Hotel Domain)},
        colback = cBlue_1!10, colframe = cBlue_7,  coltitle=white, fonttitle=\bfseries\footnotesize,
        center title, fontupper=\footnotesize, fontlower=\footnotesize]
        \textbf{Refer to the given aspect term list and select the most appropriate aspect category.} \\
        Try to match it with the aspect category list below, but if no suitable match is found, you may suggest a new category.\\
        \\
        \textbf{Aspect Category Candidates}: Cleanliness, View, Service, Facilities, Room, Pool, Parking, Breakfast, Amenity, Location, Satisfaction
    \end{tcolorbox}
    \captionof{figure}{Hotel Aspect Categorization Prompts}
    \label{fig:aspect_category_prompt}
\end{center}
\section{Experimental Result}
\label{sec:5}
\setlength{\parindent}{15pt}

This section comprehensively evaluates the performance and practicality of the proposed \muscad~framework through a series of experiments.  
First, we empirically determined the optimal number of clusters for each domain during the construction of the automatically labeled dataset using the framework. The number of aspect categories per domain was also determined through expert review.  
Then, we conducted multi-label classification experiments using the constructed datasets to compare the performance of fine-tuned classification models with that of LLM–based labeling.  
In addition, a Human Evaluation was conducted to assess whether the automatically generated labels are semantically valid in real-world domain settings.
To further verify the effectiveness of \muscad’s aspect representation learning module, we conducted comparative experiments with representative unsupervised aspect extraction methods.
The comparison was performed using topic coherence (NPMI, UMass), diversity, and embedding coherence metrics to comprehensively evaluate the quality of unsupervised aspect extraction results.
Through these experiments, we verified the reliability, scalability, and overall quality of the proposed framework.


\subsection{Experimental Settings}

In the automatic labeling process, which is one of the core components of the proposed framework, the optimal number of clusters and word extraction criteria were determined based on the characteristics of each domain. Details of the constructed datasets and the corresponding settings are summarized in Table~\ref{tab:dataset_statistics}.

\begin{table*}[htbp]
\centering
\footnotesize
\begin{tabular}{lccc}
\noalign{\hrule height 1.3pt}
\toprule
\textbf{Domain} & \textbf{Number of Clusters} & \textbf{Number of Terms} & \textbf{Final Review Count} \\
\midrule
Beauty & 13 & 150 & 255,110 \\
Hotel  & 15 & 150 & 263,395 \\
Food   & 17 & 150 & 210,272 \\
\bottomrule
\noalign{\hrule height 1.3pt}
\end{tabular}
\caption{Domain-specific Clustering and Auto-labeled Dataset Statistics}
\label{tab:dataset_statistics}
\end{table*}
The number of clusters and the number of extracted terms per cluster were determined through experiments with various hyperparameter settings. Specifically, we varied the number of extracted terms from 50 to 250 in increments of 50 to identify the optimal configuration. Detailed experimental results on the number of terms per aspect cluster are provided in \ref{sec:appendix_wordnum}. In summary, we used 15 clusters for the hotel domain, 17 for the food domain, and 13 for the beauty domain, extracting the top 150 terms from each cluster to form the initial pool of aspect term candidates. The number of heads in the Multi-Head Attention module was set to $h=4$ to effectively capture diverse contextual information within sentences.

To validate the quality of the labeled datasets generated by the framework, we fine-tuned several pre-trained language models and evaluated their classification performance. We randomly undersampled 40,000 samples for each domain. For the Korean datasets (Hotel and Food), we used KoELECTRA~\citep{park2020koelectra}, KR-BERT~\citep{lee2020krbert}, KLUE-BERT~\citep{park2021klue}, Multilingual BERT~\citep{pires-etal-2019-multilingual}, and XLM-RoBERTa~\citep{goyal-etal-2021-larger}. For the English dataset (Beauty), we used Multilingual BERT, RoBERTa~\citep{liu2020roberta}, BERT-base, BERT-large~\citep{devlin-etal-2019-bert}, XLM-RoBERTa, and DeBERTa-base~\citep{he2021deberta}.
All pre-trained language models were fine-tuned using the same hyperparameter settings: a batch size of 16, a maximum token length of 128, and a learning rate of 5e-5. We used the AdamW optimizer and trained each model for 5 epochs. During training, we applied 5-fold cross-validation to ensure stable performance across the dataset and evaluated the models using both Micro-F1 and Macro-F1 scores.

In addition, to compare the performance of fine-tuned models with large language models (LLMs), we conducted multi-aspect labeling using \texttt{GPT-3.5-turbo} and \texttt{GPT-4o-mini}. Few-shot learning was applied in the LLM experiments, and due to API cost constraints, we randomly undersampled 1,000 samples per domain for evaluation. The fine-tuned models for comparison were trained and evaluated on the same datasets using 5-fold cross-validation. For the LLM experiments, batch sizes ranged from 8 to 16, the maximum token length was set to 128, and the learning rate was set between 1e-5 and 3e-5. All experiments were conducted using a \texttt{NVIDIA GeForce RTX 3090 8GPU} environment.

\subsection{Finalization of Aspect Categories}
The aspect category lists automatically recommended in Section~\ref{sec:framework_structure} were reviewed and refined by domain experts, resulting in the final categories shown in Table~\ref{tbl2}.
\begin{table*}[h]
\centering
\footnotesize
\renewcommand{\arraystretch}{1.15}
\begin{tabular}{p{0.12\textwidth} p{0.08\textwidth} p{0.74\textwidth}}
\toprule
\textbf{Domain} & \textbf{Count} & \textbf{Aspect Category} \\
\midrule
Beauty & 10 & Satisfaction, Improvement, Usage method, Scent, Color, Persistence, Purchase, Packaging, Ingredients, Hair \\
\cmidrule(lr){1-3}
Hotel & 7 & 룸, 서비스, 만족도, 목적, 위치, 뷰, 부대시설 {\footnotesize (\textit{Translation: Room, Service, Satisfaction, Purpose, Location, View, Facilities})} \\
\cmidrule(lr){1-3}
Food & 9 & 음식, 음식량, 맛, 만족도, 위치, 분위기, 서비스, 목적, 대기시간 {\footnotesize (\textit{Translation: Food, Food Quantity, Taste, Satisfaction, Location, Atmosphere, Service, Purpose, Waiting Time})} \\
\bottomrule
\end{tabular}
\caption{Unsupervised Aspect Category Lists for Each Domain}
\label{tbl2}
\end{table*}

As shown in Table~\ref{tbl2}, the final number of aspect categories was determined to be 7 for the hotel domain, 9 for the food domain, and 10 for the beauty domain.

\subsection{Multi-Label Annotation of Review Data}

\begin{figure}
    \centering
    \includegraphics[width=\textwidth]{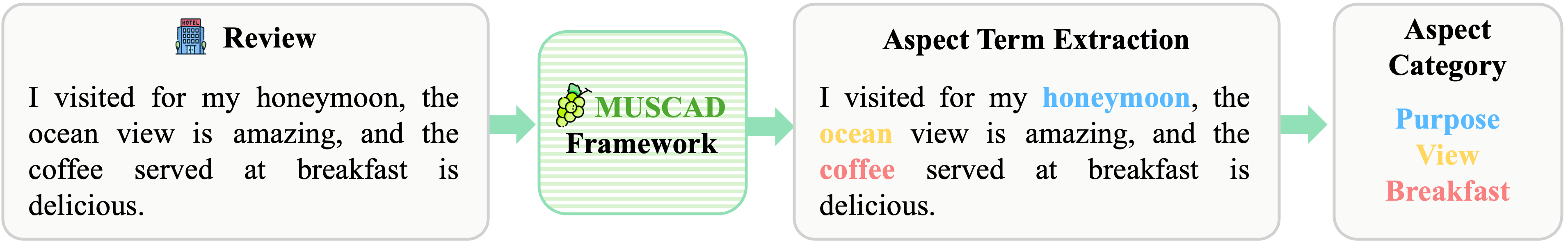}
    \caption{Pipeline of \muscad~: Aspect Term Extraction and Category Labeling from Reviews}
    \label{fig:framework_overview}
\end{figure}

Fig.~\ref{fig:framework_overview} illustrates the overall overview of the proposed \muscad~ framework.

The proposed framework processes input reviews to perform optimal clustering and automatic labeling of aspect terms and aspect categories. Using the constructed dictionaries of aspect categories and aspect terms, we conducted multi-label annotation on new review data. Considering that a single review may correspond to multiple aspect categories, the framework is designed to assign multiple labels to each review. For example, Fig.~\ref{fig:framework_overview} illustrates that a single review can be assigned to three aspect categories: ``Purpose,'' ``View,'' and ``Breakfast''.
As a result, the automatically labeled dataset consists of 255,110 reviews for the beauty domain, 263,395 for the hotel domain, and 210,272 for the food domain (see Table~\ref{tab:dataset_statistics}). This multi-labeled dataset clearly classifies the aspects addressed in each review and serves as a valuable resource for future research such as Aspect-Based Sentiment Analysis and recommendation system development. Examples of the data labeled with multiple aspects are presented in ~\ref{multi-aspect_category_labeling_data}.

\subsection{Multi-label Classification Results}

The Micro-F1 and Macro-F1 scores for each domain are presented in Table~\ref{tbl3} and Table~\ref{tbl4}, respectively.
\begin{table}[htbp]
\centering
\renewcommand{\arraystretch}{1.2}
\footnotesize
\begin{tabular}{lcc}
\noalign{\hrule height 1.3pt}
\toprule
\textbf{Model} & \textbf{Micro-F1} & \textbf{Macro-F1} \\
\midrule
BERT-base         & 97.1 & 96.3 \\
BERT-large        & 97.1 & 96.7 \\
RoBERTa           & 98.2 & 96.0 \\
DeBERta-Base      & 97.3 & 97.2 \\
Multilingual BERT & 98.0 & \textbf{97.5} \\
XLM-RoBERTa       & \textbf{98.4} & 97.4 \\
\bottomrule
\noalign{\hrule height 1.3pt}
\end{tabular}
\caption{Multi-Label Classification Results on the English Beauty Aspect-Labeled Dataset}
\label{tbl3}
\end{table}

\begin{table*}[htbp]
\centering
\renewcommand{\arraystretch}{1.2}
\footnotesize
\begin{tabular}{lcccc}
\noalign{\hrule height 1.3pt}
\toprule
\textbf{Model} & \multicolumn{2}{c}{\textbf{Hotel}} & \multicolumn{2}{c}{\textbf{Food}} \\
\cmidrule(lr){2-3} \cmidrule(lr){4-5}
 & \textbf{Micro-F1} & \textbf{Macro-F1} & \textbf{Micro-F1} & \textbf{Macro-F1} \\
\midrule
KLUE-BERT         & 98.2 & 98.1 & 97.6 & 97.2 \\
KoELECTRA         & 97.5 & 97.3 & 96.7 & 96.5 \\
KR-BERT           & 98.3 & \textbf{98.3} & \textbf{98.5} & \textbf{98.3} \\
Multilingual BERT & 98.0 & 97.5 & 96.6 & 96.4 \\
XLM-RoBERTa       & \textbf{98.4} & 98.1 & 97.3 & 97.1 \\
\bottomrule
\noalign{\hrule height 1.3pt}
\end{tabular}
\caption{Multi-Label Classification Results on the Korean Hotel \& Food Aspect-Labeled Dataset}
\label{tbl4}
\end{table*}
In the beauty dataset, Multilingual BERT and XLM-RoBERTa demonstrated the best performance. In particular, XLM-RoBERTa achieved the highest Micro-F1 score of 98.4, while Multilingual BERT recorded a Macro-F1 score of 97.5, indicating relatively balanced performance.
In the hotel and food datasets, KR-BERT and XLM-RoBERTa showed strong performance. For the hotel dataset, KR-BERT achieved the highest Macro-F1 score of 98.3, whereas XLM-RoBERTa recorded the highest Micro-F1 score of 98.4. In the food dataset, KR-BERT achieved the highest performance, with a Micro-F1 score of 98.5 and a Macro-F1 score of 98.3.
Overall, the experimental results confirm that various pre-trained language models perform well on the multi-aspect labeling dataset. In particular, XLM-RoBERTa consistently achieved high performance across all datasets, while KR-BERT demonstrated strong competitiveness in the Korean datasets.
These results verify that the labels automatically generated by the framework are of sufficient quality for training supervised models and that various pre-trained language models can achieve excellent performance on the proposed dataset.

\subsection{Comparison of Multi-Label Classification and LLM Results}
\label{llm_compare}

To evaluate the performance of \muscad, we conducted comparative experiments with large language models (LLMs).
Recent NLP research has emphasized the contrast between domain-specialized fine-tuned models and general-purpose LLMs, showing that fine-tuned models often achieve higher robustness and task-specific accuracy~\citep{QIU2024200308}.
Building on this line of research, our experiments were designed to assess whether this trend also holds in the context of multi-aspect labeling and to empirically validate the performance of the proposed \muscad~framework against LLM-based labeling approaches.
Specifically, we compared models fine-tuned on the automatically labeled datasets generated by the proposed \muscad~framework with prompted LLMs, analyzing the consistency and accuracy of each approach across multiple domains.
We extracted 1,000 samples per domain for evaluation.
Table~\ref{tbl5} and Table~\ref{tbl6} compare the results of multi-aspect labeling performed by fine-tuned models and LLMs (\texttt{GPT-3.5-turbo} and \texttt{GPT-4o-mini}) for the beauty, hotel, and food domains, respectively.
The LLM-based labeling was conducted using prompts under zero-shot and few-shot settings, and the prompts are provided in~\ref{apd:prompt}.
\begin{table}[htbp]
\centering
\renewcommand{\arraystretch}{1.2}
\footnotesize
\begin{tabular}{p{5.5cm}cc}
\noalign{\hrule height 1.3pt}
\toprule
\textbf{Model} & \textbf{Micro-F1} & \textbf{Macro-F1} \\
\midrule
\multicolumn{3}{l}{\textbf{Fine-tuned (after \muscad)}} \\
BERT-base         & 74.3 & 53.0 \\
BERT-large        & \textbf{77.5} & \textbf{64.9} \\
RoBERTa           & 75.0 & 58.0 \\
DeBERta-Base      & 66.4 & 43.4 \\
Multilingual BERT & 70.9 & 49.9 \\
XLM-RoBERTa       & 71.4 & 47.1 \\
\midrule
\multicolumn{3}{l}{\textbf{LLM (Zero-shot)}} \\
ChatGPT (GPT-3.5-turbo) & 63.6 & 54.4 \\
GPT-4o-mini                  & 52.3 & 49.7 \\
\midrule
\multicolumn{3}{l}{\textbf{LLM (Few-shot)}} \\
ChatGPT (GPT-3.5-turbo) & 57.2 & 58.9 \\
GPT-4o-mini                  & 55.7 & 49.4 \\
\bottomrule
\noalign{\hrule height 1.3pt}
\end{tabular}
\caption{Comparison Between Multi-Label Classification and LLM Results on the English Beauty Aspect-Labeled Dataset, evaluated on a test set of 1,000 samples}
\label{tbl5}
\end{table}

\begin{table}[htbp]
\centering
\renewcommand{\arraystretch}{1.2}
\footnotesize
\begin{tabular}{p{5.5cm} c c c c}
\noalign{\hrule height 1.3pt}
\toprule
\textbf{Model} 
& \multicolumn{2}{c}{\textbf{Hotel}} 
& \multicolumn{2}{c}{\textbf{Food}} \\
\cmidrule(lr){2-3}\cmidrule(lr){4-5}
& \textbf{Micro-F1} & \textbf{Macro-F1} 
& \textbf{Micro-F1} & \textbf{Macro-F1} \\
\midrule
\multicolumn{5}{l}{\textbf{Fine-tuned (after \muscad)}} \\
KLUE-BERT         & 86.0 & \textbf{85.4} & \textbf{76.1} & \textbf{75.6} \\
KoELECTRA         & 76.7 & 67.1 & 61.3 & 60.2 \\
KR-BERT           & \textbf{87.2} & 84.8 & 74.1 & 72.2 \\
Multilingual BERT & 81.9 & 78.2 & 55.8 & 50.5 \\
XLM-RoBERTa       & 61.6 & 48.8 & 71.1 & 69.2 \\
\midrule
\multicolumn{5}{l}{\textbf{LLM (Zero-shot)}} \\
ChatGPT (GPT-3.5-turbo) & 71.1 & 70.0 & 62.3 & 50.5 \\
GPT-4o-mini             & 73.2 & 70.5 & 57.4 & 47.7 \\
\midrule
\multicolumn{5}{l}{\textbf{LLM (Few-shot)}} \\
ChatGPT (GPT-3.5-turbo) & 73.2 & 65.4 & 61.0 & 50.7 \\
GPT-4o-mini             & 76.0 & 72.7 & 61.1 & 48.6 \\
\bottomrule
\noalign{\hrule height 1.3pt}
\end{tabular}
\caption{Comparison Between Multi-Label Classification and LLM Results on the Korean Hotel \& Food Aspect-Labeled Dataset, evaluated on text set of 1,000 samples}
\label{tbl6}
\end{table}
The comparison results can be summarized in three key findings. \\\
\textbf{Performance}~~Fine-tuned models, which perform labeling based on data trained for specific domains, generally achieved higher Micro-F1 and Macro-F1 scores. In particular, KR-BERT and XLM-RoBERTa showed the best performance in the hotel and food domains, while BERT-large and RoBERTa performed relatively well in the beauty domain. On the other hand, the labeling results using LLMs were generally inferior to those of the fine-tuned models. Although the LLMs demonstrated the ability to interpret sentence meaning and assign appropriate aspects, issues were observed such as assigning incorrect labels or producing inconsistent results for the same sentence (see Table~\ref{tbl:label_comparison}).
\begin{table*}[htbp]
\centering
\renewcommand{\arraystretch}{1.2}
\footnotesize
\begin{tabular}{p{8.6cm} p{2.5cm} p{3.5cm}}
\noalign{\hrule height 1.3pt}
\toprule
\textbf{Review Text} & \textbf{\muscad~} & \textbf{LLM-Based Labels} \\
\midrule
\shortstack[l]{Two men traveled, and perhaps because it was \\off-season,
there were hardly any people; it was quiet \\and offered good value for the price.}
& \shortstack[l]{Satisfaction \cmark \\ Purpose \cmark}
& \shortstack[l]{Satisfaction \cmark \\ Location \xmark} \\ 
\midrule
\shortstack[l]{The tangsuyuk is also tasty, but since it's located on the\\ first floor of a building where many companies are \\ situated, there's a really long wait at lunchtime.}
& \shortstack[l]{Food \cmark \\ Satisfaction \cmark \\ Waiting Time \cmark}
& \shortstack[l]{Taste \xmark \\ Location \xmark \\ Waiting Time \cmark} \\
\midrule
\shortstack[l]{I returned this item because it had terrible coverage,\\similar to using a skin tone primer}
& \shortstack[l]{Color \cmark \\ Satisfaction \cmark}
& \shortstack[l]{() \xmark \\ Satisfaction \cmark } \\
\bottomrule
\noalign{\hrule height 1.3pt}
\end{tabular}
\caption{Comparison of Aspect Category Labeling by \muscad~and an LLM-Based Model. 
A green check (\cmark) indicates a correctly identified aspect, whereas a red cross (\xmark) denotes an incorrect or missing aspect.}
\label{tbl:label_comparison}
\end{table*}
As shown in Table~\ref{tbl:label_comparison}, for the same review, the fine-tuned model correctly predicted the aspects ``Satisfaction'' and ``Purpose'' whereas the LLM assigned ``Satisfaction'' and ``Location'' resulting in partially inaccurate labeling. This suggests that while LLMs are capable of interpreting context and extracting appropriate aspects, they tend to assign semantically inappropriate labels in certain contexts. \\
\textbf{Reproducibility}~~LLM-based labeling exhibited issues with reproducibility. When the same review was input multiple times, the resulting labels were often inconsistent, especially in few-shot settings. This indicates that the outputs of LLMs can vary significantly depending on prompt engineering and the structure of input examples, suggesting that fine-tuned models may be a more reliable alternative for tasks where consistency is critical, such as large-scale data labeling. \\
\textbf{Cost}~~Multi-aspect labeling using LLMs has the advantage of being applicable to various domains without additional data training. However, when processing large volumes of data, the cost becomes a significant drawback. With models such as \texttt{GPT-4o-mini} and \texttt{GPT-3.5-turbo}, API call costs accumulate rapidly, and labeling hundreds of thousands of reviews can lead to substantial expenses.  
In contrast, fine-tuned models require initial training costs, but once trained, they can efficiently process large datasets without additional costs. Moreover, they can be run on local servers, enabling real-time labeling without delays from API calls. In this regard, fine-tuned models offer better cost-efficiency and scalability compared to LLMs.

In conclusion, while LLM-based multi-aspect labeling can be useful for quickly labeling small datasets, fine-tuned models prove to be a more practical solution for large-scale data processing. Moreover, these results are consistent with recent studies that highlight the instruction-following limitations of large language models~\citep{zan-etal-2025-building}.
That work points out the fundamental limitation of LLMs in strictly adhering to complex task instructions and suggests that additional fine-tuning approaches are required to mitigate this issue.
Our study also observed similar limitations while LLMs exhibit strong contextual understanding, they still show weaknesses in terms of output stability and reproducibility. They demonstrate higher reproducibility and more consistent performance compared to LLMs, and are more cost-effective, making them better suited for real-world applications.

\subsection{Human Evaluation}

To qualitatively evaluate the quality of the automated \muscad~framework, we conducted a Human Evaluation. To assess whether the labels generated by \muscad~are semantically appropriate in practice, three domain experts were recruited to manually perform multi-aspect category labeling on 100 review samples from each domain. Clear guidelines explaining each aspect category and the labeling criteria were provided to the evaluators to ensure consistency, and the guideline is illustrated in Fig.~\ref{fig:human_evaluation_guide}.
\begin{tcolorbox}
    [title={Human Evaluation Criteria},
    colback = green!5, colframe = green,  coltitle=black,fonttitle=\bfseries\footnotesize,
    center title,fontupper=\footnotesize,fontlower=\footnotesize]
    \footnotesize
    \textbf{This survey is a Human Evaluation task that involves selecting appropriate labels based on review data.} Please read the given reviews carefully and choose the most relevant label (s). Each item focuses on describing a specific aspect of the review. The reviews are from Amazon’s Beauty \& Personal Care product category.  
    The labeling method and guidelines are well explained below, so please read them thoroughly before proceeding.

    \textbf{Labeling Method:}
    \begin{itemize}
        \item Select the most relevant label (s) based on the content of the review.
        \item If multiple labels apply, check all applicable labels.
    \end{itemize}

    \vspace{2mm}
    \textbf{Labeling Guidelines:}
    \begin{itemize}
        \item \textbf{Satisfaction:} Expressions related to overall satisfaction with the experience.  
        \item \textbf{Ingredients:} Mentions of product ingredients, raw materials, or composition.  
        \item \textbf{Usage Method:} References to product usage methods or applications (e.g., face, body, cosmetics, etc.).  
        \item \textbf{Improvement:} Aspects that have improved after using the product.  
        \item \textbf{Color:} Opinions related to the product’s color.  
        \item \textbf{Hair:} Effects on hair after using the product or changes related to hair condition.  
        \item \textbf{Packaging:} Mentions of packaging design, durability, shipping condition, etc.  
        \item \textbf{Scent:} Descriptions of fragrance, intensity, longevity, or types of scent.  
        \item \textbf{Purchase:} Mentions of purchasing experience, reasons for purchase, or target audience for the product.  
    \end{itemize}
    \end{tcolorbox}
\captionof{figure}{Human Evaluation Criteria for Qualitative Assessment}
\label{fig:human_evaluation_guide}
The evaluation was conducted by comparing the labels generated by \muscad~with those assigned by each evaluator using Micro-F1 and Macro-F1 scores. The results are presented in Table~\ref{tbl7}.
\begin{table}[htbp]
\centering
\renewcommand{\arraystretch}{1.2}
\footnotesize
\begin{tabular}{lcccccc}
\noalign{\hrule height 1.3pt}
\toprule
\textbf{Evaluator} 
& \multicolumn{2}{c}{\textbf{Hotel}} 
& \multicolumn{2}{c}{\textbf{Food}} 
& \multicolumn{2}{c}{\textbf{Beauty}} \\
\cmidrule(lr){2-3}\cmidrule(lr){4-5}\cmidrule(lr){6-7}
& \textbf{Micro-F1} & \textbf{Macro-F1}
& \textbf{Micro-F1} & \textbf{Macro-F1}
& \textbf{Micro-F1} & \textbf{Macro-F1} \\
\midrule
Evaluator 1 & 78.6 & 79.3 & 60.1 & 61.1 & 67.7 & 61.2 \\
Evaluator 2 & 69.4 & 72.5 & 74.1 & 71.7 & 66.7 & 61.6 \\
Evaluator 3 & 69.1 & 72.1 & 75.8 & 72.9 & 69.7 & 61.6 \\
\midrule
\textbf{Mean (Std)} & 72.4 \footnotesize{(4.4)} & 74.6 \footnotesize{(3.4)} & 70.0 \footnotesize{(7.0)} & 68.6 \footnotesize{(5.4)} & 68.0 \footnotesize{(1.2)} & 61.5 \footnotesize{(0.2)} \\
\bottomrule
\noalign{\hrule height 1.3pt}
\end{tabular}
\caption{Human Evaluation Results with Average, Standard Deviation, and \muscad~Comparison}
\label{tbl7}
\end{table}
According to the Human Evaluation results presented in Table~\ref{tbl7}, the labels automatically generated by \muscad~maintained a consistent level of performance when compared to those assigned by human evaluators. This indicates that the proposed framework performs aspect extraction and classification tasks at a reliable level within specific domains.
The Human Evaluation also revealed differences among evaluators, with some domains showing inconsistent application of labeling criteria. This suggests that human interpretation may vary depending on the individual, and that a clear understanding of the guidelines is essential for maintaining consistency in labeling specific aspect categories. In particular, there were cases where the same review was labeled differently by different evaluators, highlighting the potential for subjectivity in manual labeling.
In contrast, \muscad~applies automated labeling based on a consistent standard, which minimizes variability among evaluators and produces more uniform results. Manual labeling requires expert involvement and is both time-consuming and costly. Even when guidelines are provided, differences in interpretation can arise. However, automated labeling using the proposed framework overcomes these limitations by enabling fast and consistent processing of large-scale data.

\setlength{\parindent}{15pt}
\subsection{Comparison with Unsupervised Aspect Extraction Methods}
\begin{table}[htbp]
\centering
\renewcommand{\arraystretch}{1.2}
\footnotesize
\begin{tabular}{lllccccc}
\noalign{\hrule height 1.3pt}
\textbf{Domain} & \textbf{Model Type} & \textbf{Model} & \textbf{NPMI} & \textbf{UMass} & \textbf{Diversity} & \textbf{EC} & \textbf{Rank Avg} \\ 
\noalign{\hrule height 1.3pt}
\multirow{10}{*}{Hotel} & \multirow{3}{*}{Probabilistic-based} & \texttt{LDA}      & 0.026    & -3.833   & \textbf{1.000}   & –     & 5.7  \\
                        &                                      & \texttt{BTM}      & -0.337   & -18.132  & 0.143   & –     & 9.7  \\
                        &                                      & \texttt{NMF}      & 0.038    & -3.770   & 0.671   & –     & 7.0  \\ \cmidrule(lr){2-8}
                        & Hybrid                               & \texttt{CTM}      & 0.195    & -0.229   & \underline{0.871}   & 0.499 & 3.3  \\ \cmidrule(lr){2-8}
                        & \multirow{6}{*}{Embedding-based}     & \texttt{BERTopic} & 0.173    & -0.480   & 0.692   & \underline{0.550} & 4.3  \\
                        &                                      & \texttt{Top2Vec}  & 0.025    & -0.683   & 0.094   & 0.502 & 7.3  \\
                        &                                      & \texttt{KeyBERT}  & 0.087    & -0.382   & 0.707   & 0.266 & 5.3  \\
                        &                                      & \texttt{SimCSE}   & -0.018   & -0.534   & 0.343   & 0.520 & 6.5  \\
                        &                                      & \texttt{ABAE}     & \underline{0.242}    & \underline{-0.266}   & 0.863   & 0.486 & \underline{3.8}  \\
                        &                                      & \texttt{MUSCAD}   & \textbf{0.281}    & \textbf{-0.220}   & \textbf{1.000 }  & \textbf{0.568} & \textbf{1.0}  \\ \midrule
\multirow{10}{*}{Food}  & \multirow{3}{*}{Probabilistic-based} & \texttt{LDA}      & 0.021    & -4.302   & \textbf{1.000}   & –     & 5.0  \\
                        &                                      & \texttt{BTM}      & -0.459   & -20.227  & 0.111   & –     & 10.0 \\
                        &                                      & \texttt{NMF}      & 0.017    & -3.994   & 0.656   & –     & 6.7  \\ \cmidrule(lr){2-8}
                        & Hybrid                               & \texttt{CTM}      & 0.222    & -0.279   & 0.715   & 0.500 & 4.3  \\ \cmidrule(lr){2-8}
                        & \multirow{6}{*}{Embedding-based}     & \texttt{BERTopic} & 0.189    & -0.409   & 0.528   & \underline{0.561} & 4.3  \\
                        &                                      & \texttt{Top2Vec}  & 0.021    & -0.592   & 0.098   & 0.510 & 6.0  \\
                        &                                      & \texttt{KeyBERT}  & 0.078    & -0.630   & 0.705   & 0.508 & 5.0  \\
                        &                                      & \texttt{SimCSE}   & -0.015   & -0.763   & 0.310   & 0.527 & 7.3  \\
                        &                                      & \texttt{ABAE}     & \underline{0.334}    & \underline{-0.052}   & \underline{0.991}   & 0.491 & \underline{3.3}  \\
                        &                                      & \texttt{MUSCAD}   & \textbf{0.384}    & \textbf{-0.050}   & \textbf{1.000}   & \textbf{0.592} & \textbf{1.0}  \\ \midrule
\multirow{10}{*}{Beauty} & \multirow{3}{*}{Probabilistic-based} & \texttt{LDA}      & 0.017    & -4.416   & 1.000   & –     & 5.7  \\
                        &                                      & \texttt{BTM}      & -0.493   & -20.837  & 0.100   & –     & 10.0 \\
                        &                                      & \texttt{NMF}      & 0.018    & -3.923   & 0.600   & –     & 7.0  \\ \cmidrule(lr){2-8}
                        & Hybrid                               & \texttt{CTM}      & 0.267    & -0.123   & \underline{0.939}   & 0.281 & 3.8  \\ \cmidrule(lr){2-8}
                        & \multirow{6}{*}{Embedding-based}     & \texttt{BERTopic} & 0.146    & -0.357   & 0.769   & 0.292 & 4.8  \\
                        &                                      & \texttt{Top2Vec}  & 0.102    & -0.918   & 0.161   & 0.325 & 6.0  \\
                        &                                      & \texttt{KeyBERT}  & 0.170    & -0.521   & 0.856   & 0.279 & 5.0  \\
                        &                                      & \texttt{SimCSE}   & -0.008   & -0.667   & 0.444   & \underline{0.332} & 7.0  \\
                        &                                      & \texttt{ABAE}     & \underline{0.313}    & \underline{-0.016}   & \underline{1.000}   & 0.274 & \underline{3.0}  \\
                        &                                      & \texttt{MUSCAD}   & \textbf{0.384}    & \textbf{-0.008}   & \textbf{1.000}   & \textbf{0.360} & \textbf{1.0}  \\ 
\noalign{\hrule height 1.3pt}
\end{tabular}
\caption{Comparison of Topic Modeling Methods across Domains}
\label{tbl:comparison_aspect}
\end{table}

To objectively evaluate the quality of automatically extracted Aspect Terms,
it is essential to employ evaluation metrics that can measure both the semantic coherence among Aspect Terms and the distinctiveness between different Aspect Categories, even in the absence of labeled data.
\citep{NPMI},~\citep{UMass} proposed coherence- and diversity-based evaluation metrics that have been widely adopted as quantitative measures showing strong correlations with human judgments of topic interpretability.
Building upon these approaches, we comprehensively evaluate the performance of~\muscad~using four complementary metrics: NPMI, UMass, Diversity, and Embedding Coherence.
NPMI and UMass assess the semantic and document-level coherence among top-ranked Aspect Terms,
Diversity measures the redundancy across Aspect Terms,
and Embedding Coherence evaluates the degree of semantic alignment in the embedding space.
Together, these four metrics provide complementary perspectives that enable a holistic evaluation of how effectively~\muscad~learns coherent and well-separated Aspect representations in an unsupervised setting.

As introduced above, the topic coherence metrics, NPMI and UMass, are used to quantitatively evaluate the semantic consistency among the top-ranked words within each Aspect. In this study, both NPMI and UMass metrics were employed to achieve a balanced evaluation of semantic relatedness and document-level co-occurrence consistency.
The Pointwise Mutual Information (PMI) measures the association strength between two words based on the ratio of their joint probability $P(w_i, w_j)$ to the product of their individual probabilities $P(w_i)$ and $P(w_j)$ .
A small smoothing constant $\epsilon = 1$ is applied to prevent zero probabilities during the logarithmic operation.
\begin{equation}
    \text{PMI}(w_i, w_j) = \log_2 \left( \frac{P(w_i, w_j) + \epsilon}{P(w_i)P(w_j)} \right)
    \label{eq7}
\end{equation}
The Normalized Pointwise Mutual Information (NPMI) is widely used because it shows a strong correlation with human judgments of topic interpretability~\citep{NPMI}.
NPMI extends PMI by normalizing its value to the range $[-1, 1]$, providing a more interpretable scale for assessing the strength of word associations.
The NPMI between two words $w_i$ and $w_j$ is defined as follows:
\begin{equation}
    \text{NPMI}(w_i, w_j) =
    \frac{\log \frac{P(w_i, w_j) + \epsilon}{P(w_i) P(w_j)}}
    {-\log (P(w_i, w_j) + \epsilon)}
\label{eq8}
\end{equation}
Here, $P(w_i, w_j)$ denotes the probability that the two words co-occur within a sliding window,
and $\epsilon$ is a smoothing constant used to avoid zero probabilities.
A higher NPMI value indicates stronger semantic associations among Aspect Terms,
implying greater intra-aspect cohesion.

UMass Coherence evaluates topic quality based on the co-occurrence of top-ranked words within the same document~\citep{UMass}.
While NPMI relies on mutual information, UMass measures the average pairwise log conditional probability between words. It is defined as follows:
\begin{equation}
    \text{UMass}(w_i, w_j) =
    \log \frac{D(w_i, w_j) + \epsilon}{D(w_j)}
\label{eq9}
\end{equation}
Here, $D(w_i, w_j)$ denotes the number of documents in which the two words co-occur, and $D(w_j)$ represents the number of documents containing the word $w_j$. The small constant $\epsilon$ is added to avoid division by zero.
A UMass value closer to zero indicates higher topic coherence.

Diversity measures the degree of redundancy among aspect categories.
It is computed as the ratio of unique words to the total number of words across the top $N$ terms from all aspects:
\begin{equation}
    \text{Diversity} =
    \frac{\text{Number of unique words among top } N \text{ terms}}{N \times K}
    \label{eq10}
\end{equation}
A higher Diversity score indicates that each aspect captures distinct and non-overlapping semantic concepts.
This suggests that the learned aspect embeddings are well separated in the semantic space.

Embedding Coherence (EC) complements traditional co-occurrence-based metrics by measuring the semantic similarity among top aspect terms in the embedding space.
It is defined as the average pairwise cosine similarity among the embeddings of the top $N$ aspect terms:
\begin{equation}
    \text{EC}(W) =
    \frac{1}{\binom{N}{2}}
    \sum_{i=1}^{N-1} \sum_{j=i+1}^{N}
    \cos(\text{Vec}(w_i), \text{Vec}(w_j))
    \label{eq11}
\end{equation}
A higher EC value indicates that the embeddings of words within the same aspect category are semantically closer to each other.
Unlike simple co-occurrence statistics, this metric reflects not only the frequency of word co-occurrence but also their semantic orientation and contextual similarity in the embedding space, providing a more comprehensive measure of how coherently the model captures semantic concepts.

To verify the effectiveness of the unsupervised aspect representation learned by~\muscad, we conducted comparative experiments with nine representative unsupervised aspect extraction methods: \texttt{LDA}, \texttt{BTM}, \texttt{NMF}, \texttt{BERTopic}, \texttt{Top2Vec}, \texttt{CTM}, \texttt{KeyBERT}, \texttt{SimCSE}, and \texttt{ABAE}.
Further details and implementation notes for the nine comparative models can be found in~\ref{ae_method_details}.
All models were configured with the same number of aspect categories and aspect terms to ensure a fair comparison.
In addition to quantitative evaluation, we also visualized the top keywords extracted by each model and performed qualitative comparative analysis to assess both the interpretability and semantic coherence of the aspects generated by each model.

Table~\ref{tbl:comparison_aspect} demonstrates that~\muscad~outperforms existing unsupervised learning-based methods across all three domains (Hotel, Food, and Beauty).
~\muscad~achieves the highest scores in NPMI, Diversity, and Embedding Coherence, while also maintaining stable UMass values close to zero.
These results indicate that~\muscad~effectively learns aspect representations that are semantically coherent, non-redundant, and contextually well-aligned.
In addition, the average rank (Rank Avg) is the lowest across all domains, confirming the overall balance and robustness of the model.
These findings show that~\muscad~exhibits higher interpretability than probabilistic models and greater contextual coherence than embedding-based models.

To gain deeper qualitative insights into the differences among models, we compared the top 20 Aspect Terms belonging to the scent Aspect Category in the Beauty domain.
As shown in Table~\ref{tbl:comparison_aspect_term_beauty}, traditional probabilistic models such as \texttt{LDA} and \texttt{BTM} tend to generate generic and loosely related words (e.g., smell, scent, fragrance, soap, lotion) rather than domain-specific fragrance terms.
Embedding-based approaches such as \texttt{BERTopic} and \texttt{KeyBERT} capture a broader contextual range but often mix semantically irrelevant attributes (e.g., hair, body, red, clean), leading to reduced coherence within the aspect cluster.

In contrast, \muscad~produces a more semantically cohesive and domain-reflective set of terms, including vanilla, lavender, sandalwood, jasmine, and cedarwood, which capture fine-grained nuances of fragrance-related semantics.
These findings demonstrate that \muscad~can consistently learn coherent and semantically aligned Aspect representations across domains in an unsupervised setting, while generating interpretable and fine-grained Aspect structures that align well with human perception.

\begin{table}[H]
\centering
\renewcommand{\arraystretch}{1.2} 
\footnotesize
\begin{tabular}{p{2cm} p{13cm}}
\noalign{\hrule height 1.3pt}
\toprule
\textbf{Model} & \textbf{Aspect Term} \\ 
\midrule
\textbf{\texttt{LDA}} & smell, scent, perfume, pleasant, fresh, soap, clean, shower, body, wash, creamy, feel, nice, water, mild, chemical, moisturize, absorbed, soft, clear \\  
\cmidrule(lr){1-2}
\textbf{\texttt{BTM}} & scent, buy, make, start, beauty, nice, best, careful, lip, girl, tell, show, start, say, turn, let, make, mix, husband, hair \\  
\cmidrule(lr){1-2}
\textbf{\texttt{NMF}} & smell, scent, fragrance, perfume, pleasant, fresh, sweet, soap, lotion, oil, soft, clean, creamy, smooth, silky, moisturizing, chemical, body, wash, lingers \\  
\cmidrule(lr){1-2}
\textbf{\texttt{CTM}} & lovely, smell, pleasant, scent, sweet, subtle, overpowering, nice, wonderful, nicely, creamy, smoothly, fragrance, strong, floral, medium, gorgeous, blush, lightweight, matte \\   
\cmidrule(lr){1-2}
\textbf{\texttt{BERTopic}} & hair, skin, smell, shampoo, well, cream, soft, oil, natural, every, body, red, clean, soap, iron, light, think, leaf, wear, last \\   
\cmidrule(lr){1-2}
\textbf{\texttt{Top2Vec}} & scent, fragrance, perfume, smell, soap, lotion, wash, spray, shampoo, fresh, coconut, oily, greasy, moisturizing, moisturizer, clean, shower, creamy, silky, pleasant \\
\cmidrule(lr){1-2}
\textbf{\texttt{KeyBERT}} & odor, smell, perfume, fragrance, greasy, leaf, invigorating, nice, clean, fresh, soap, body, lotion, water, skin, hair, color, fair, powdery, pleasant \\
\cmidrule(lr){1-2}
\textbf{\texttt{SimCSE}} & smell, scent, like, love, product, good, fragrance, nice, great, really, oil, feel, perfume, strong, doesnt, light, dont, pleasant, skin, little \\
\cmidrule(lr){1-2}
\textbf{\texttt{ABAE}} & aroma, mint, fruity, floral, overpowering, scent, vanilla, sweet, smell, musk, flowery, scented, feminine, smelling, pleasing, musky, flavor, masculine, hint, vibe \\
\cmidrule(lr){1-2}
\textbf{\texttt{MUSCAD}} & pungent, fruity, flowery, musky, woodsy, spicy, floral, earthy, minty, sweet, citrusy, patchouli, sandalwood, vanilla, lavender, peppermint, tabacco, bergamot, jasmine, cedarwood \\
\bottomrule
\noalign{\hrule height 1.3pt}
\end{tabular}
\caption{Comparison of Top 20 Aspect Terms for Scent Category in Beauty Domain by Different Topic Modeling Methods}
\label{tbl:comparison_aspect_term_beauty}
\end{table}

\subsection{Summary}
As demonstrated through various experiments in Section~\ref{sec:5}, the multi-aspect labels generated by the proposed framework are of sufficient quality for use in model training. The multi-label classification experiments using various fine-tuned models yielded high Micro-F1 and Macro-F1 scores, indicating that the automatically labeled data is suitable for supervised learning.
Furthermore, we compared the performance of fine-tuned models with few-shot labeling using \texttt{GPT-3.5-turbo} and \texttt{GPT-4o-mini} to evaluate the capabilities of LLMs in multi-aspect labeling. The results showed that while LLMs were able to assign appropriate labels in some cases, their overall performance was inferior to that of the fine-tuned models. LLMs also exhibited issues with reproducibility and incurred higher costs.
In addition, Human Evaluation confirmed that the automatic labels produced by \muscad~achieved a level of reliability and consistency comparable to manual labeling by experts, and were more effective in reducing subjectivity among evaluators. These findings suggest that the proposed framework offers a more practical solution for applying multi-aspect labeling to large-scale review data.
In addition to the supervised evaluation, we further validated the effectiveness of~\muscad~in learning unsupervised aspect representations.
Comparative experiments with nine representative unsupervised aspect extraction methods—\texttt{LDA}, \texttt{BTM}, \texttt{NMF}, \texttt{CTM}, \texttt{BERTopic}, \texttt{Top2Vec}, \texttt{KeyBERT}, \texttt{SimCSE}, and \texttt{ABAE}—demonstrated that~\muscad~consistently achieved the highest scores across multiple coherence and diversity metrics (NPMI, UMass, Diversity, and Embedding Coherence).
These results confirm that~\muscad~learns semantically coherent, non-redundant, and domain-reflective aspect representations, outperforming existing probabilistic and embedding-based approaches.
Taken together, the results of both the supervised and unsupervised evaluations highlight the robustness and generalizability of~\muscad~as a comprehensive framework for scalable and interpretable multi aspect representation learning.
\section{Discussion}
\subsection{Conclusion}

In this study, we proposed \muscad~, an unsupervised multi-aspect labeling framework designed to automatically assign aspect categories to multilingual and multi-domain review data.
Unlike previous studies that rely on supervised learning approaches requiring large amounts of manually annotated data, the proposed framework integrates K-means–based aspect candidate generation, contextual representation learning using Multi-Head Attention, and classification optimization through Max-Margin Loss, enabling multi-aspect labeling without predefined vocabularies.
Experimental results showed that fine-tuned classification models trained on the multi-aspect labeling dataset generated by the proposed framework achieved high F1-scores, demonstrating the reliability of the automatically generated labels.
Furthermore, compared to large language models (LLMs), \muscad~produced more consistent and reproducible labeling outcomes, while Human Evaluation confirmed that the quality of its automatically generated labels was comparable to expert annotations.
In addition, unsupervised evaluation experiments revealed that \muscad~effectively learns semantically coherent and well-separated aspect representations, outperforming representative unsupervised aspect extraction methods.
Finally, the \muscad~framework successfully performed large-scale automatic labeling across multiple languages and domains, thereby demonstrating both its practical applicability and scalability in real-world environments.

\subsection{Future Work}

This study proposed and evaluated the performance of the unsupervised \muscad~framework across diverse domains. However, several directions remain for future improvements and extensions. 
First, we aim to build a fully automated labeling system using AI agents. Currently, partial manual intervention is required for matching aspect categories. In future work, we plan to eliminate manual processes entirely by incorporating AI agents and establishing an automated feedback loop, ultimately enabling fully autonomous labeling.

Second, developing aspect-aware review summarization is another important direction. \muscad~can automatically label multiple aspects within a sentence and group sentences that share the same aspect, providing a foundation for aspect-level summarization. Furthermore, such aspect-based summaries can be extended to ``AI agent systems'' that summarize user logs or review histories to build personalized profiles. For example, aspect summaries such as ``service'', ``food quality'', and ``ambience'' generated by \muscad~can be transformed into aspect-based user profiles that reflect user preferences, which can then be utilized for personalized responses in recommendation systems or conversational agents.

Third, the multi-aspect labeled datasets automatically generated by \muscad~can be leveraged in various future research directions. Beyond serving as input for Aspect-Based Sentiment Analysis (ABSA) models, they can contribute to enhancing contextual understanding and representation learning in knowledge-based sentiment analysis models such as the Knowledge Graph Augmented Network (KGAN)~\citep{zhong2023kgan}. Moreover, due to \muscad’s scalability across multiple languages and specialized domains, it can provide a foundation for new aspect discovery research even in low-resource environments such as healthcare, legal, or manufacturing domains.

These research directions represent a path for \muscad~to evolve from a simple unsupervised labeling model into an AI agent–driven, personalization-oriented, and aspect-understanding framework. Through this evolution, \muscad~can serve as a key foundation connecting aspect-level summarization, user profiling, multilingual and domain-adaptive labeling, and future advances in Aspect-Based Sentiment Analysis (ABSA) and knowledge-based sentiment understanding.

\section*{Acknowledgements}
This work was supported by the Institute of Information \& Communications Technology Planning \& Evaluation (IITP) grant funded by the Korea government (MSIT) (RS-2024-00337072), the National Research Foundation of Korea (NRF) grant funded by the Korea government (MSIT) (No. RS2024-0040780), and the Institute of Information \& Communications Technology Planning \& Evaluation (IITP) grant funded by the Korea government (MSIT) (No. RS-2020-II201373, Artificial Intelligence Graduate School Program, Hanyang University). This research was supported by Basic Science Research Program through the
National Research Foundation of Korea(NRF) funded by the Ministry of Education(RS-2025-2548326)

\appendix

\renewcommand{\thetable}{\Alph{section}.\arabic{table}} 
\setcounter{table}{0} 

\renewcommand{\thefigure}{\Alph{section}.\arabic{figure}}
\setcounter{figure}{0}

\section{Experiment on the Number of Words per Aspect Category}
\label{sec:appendix_wordnum}

The number of words extracted for each aspect category affects both the model’s performance and the data coverage. To determine the optimal number of words, we conducted experiments under various settings to compare performance.
Specifically, we constructed multi-aspect labeling datasets by extracting the top 50, 100, 150, 200, and 250 words for each aspect category, and evaluated the performance using the XLM-RoBERTa model. The main objective of this experiment was to analyze how the number of extracted words influences classification performance (F1-score) and coverage (i.e., the proportion of relevant words included within each aspect category).
Through this experiment, we aimed to identify the optimal number of words that maximizes generalization performance while maintaining model accuracy. The results and analysis are presented in Table~\ref{tbl8} and Table~\ref{tbl9}.

\begin{table}[h]
\centering
\renewcommand{\arraystretch}{1.05}
\footnotesize
\begin{tabular}{l l l l}
\noalign{\hrule height 1.3pt}
\textbf{Word Count} & \textbf{Hotel} & \textbf{Food} & \textbf{Beauty} \\ 
\midrule
50  &  97.17\%  &  76.14\%  &  73.32\%  \\ 
100 &  98.41\% {\scriptsize(+1.24\%)} &  87.55\% {\scriptsize(+11.41\%)} &  85.43\% {\scriptsize(+12.11\%)} \\
150 &  98.60\% {\scriptsize(+0.19\%)} &  84.89\% {\scriptsize(-2.66\%)}  &  92.24\% {\scriptsize(+6.81\%)}  \\ 
200 &  98.81\% {\scriptsize(+0.21\%)} &  83.85\% {\scriptsize(-1.04\%)}  &  95.20\% {\scriptsize(+2.96\%)}  \\ 
250 &  98.96\% {\scriptsize(+0.15\%)} &  97.51\% {\scriptsize(+13.66\%)} &  96.31\% {\scriptsize(+1.11\%)}  \\ 
\noalign{\hrule height 1.3pt}
\end{tabular}
\caption{Coverage of Top-N Words across Three Domains (Hotel, Food, Beauty).
Each row corresponds to a specific number of words (e.g., 50, 100, 150, etc.),
and the table shows the cumulative percentage of tokens covered in each domain.
Values in parentheses indicate the relative change in coverage compared to the
immediately preceding row.}
\label{tbl8}
\end{table}

As shown in Table~\ref{tbl8}, increasing the number of words per aspect category from 50 to 100 led to a significant improvement in coverage across all domains. In particular, the food and beauty domains showed substantial increases of +11.41\% and +12.11\%, respectively.
At 150 words, coverage continued to improve in the hotel and beauty domains; however, a slight decrease of –2.66\% was observed in the food domain. When increasing the word count beyond 200, the gains became marginal or even negative, suggesting that including excessive terms may lead to performance degradation.
Based on these findings, and considering both coverage improvement and generalization performance, we selected 150 words as the optimal number for each aspect category.

\subsection{Performance Analysis by Number of Words}

\begin{table}[h]
\centering
\renewcommand{\arraystretch}{1.1} 
\begin{adjustbox}{center,max width=0.95\textwidth}
\footnotesize
\begin{tabular}{p{1.4cm} p{2.4cm} p{2.4cm} p{2.4cm} p{2.4cm} p{2.4cm} p{2.4cm}} 
\noalign{\hrule height 1.3pt}
\toprule
 & \multicolumn{2}{c}{\textbf{Hotel}} & \multicolumn{2}{c}{\textbf{Food}} & \multicolumn{2}{c}{\textbf{Beauty}} \\ 
\cmidrule(lr){2-3} \cmidrule(lr){4-5} \cmidrule(lr){6-7}
\textbf{Word Count} & \textbf{Micro-F1} & \textbf{Macro-F1} & \textbf{Micro-F1} & \textbf{Macro-F1} & \textbf{Micro-F1} & \textbf{Macro-F1} \\ 
\midrule
50  & 0.9785 & 0.974  & 0.9405 & 0.9022 & 0.9655 & 0.8658 \\ 
100 & 0.9731 {\scriptsize(+0.54\%)} & 0.9726 {\scriptsize(+0.14\%)} & 0.9291 {\scriptsize(-1.14\%)} & 0.9102 {\scriptsize(+0.8\%)} & 0.9477 {\scriptsize(-1.78\%)} & 0.8774 {\scriptsize(+1.16\%)} \\ 
150 & 0.9751 {\scriptsize(+0.20\%)} & 0.9732 {\scriptsize(+0.06\%)} & 0.9097 {\scriptsize(-1.95\%)} & 0.8965 {\scriptsize(-1.37\%)} & 0.9439 {\scriptsize(-0.38\%)} & 0.8841 {\scriptsize(+0.67\%)} \\ 
200 & 0.9778 {\scriptsize(+0.27\%)} & 0.9757 {\scriptsize(+0.25\%)} & 0.9051 {\scriptsize(-0.46\%)} & 0.8853 {\scriptsize(-1.12\%)} & 0.9429 {\scriptsize(-0.10\%)} & 0.8686 {\scriptsize(+1.55\%)} \\ 
250 & 0.9713 {\scriptsize(-0.65\%)} & 0.9688 {\scriptsize(-0.69\%)} & 0.9202 {\scriptsize(+1.51\%)} & 0.9066 {\scriptsize(+2.13\%)} & 0.9412 {\scriptsize(-0.17\%)} & 0.8708 {\scriptsize(+0.22\%)} \\ 
\bottomrule
\noalign{\hrule height 1.3pt}
\end{tabular}
\end{adjustbox}
\caption{Performance and Coverage of XLM-RoBERTa with Varying Top-N Word Counts. The table reports the Micro-F1 and Macro-F1 scores for Hotel, Food, and Beauty domains when selecting the top 50, 100, 150, 200, and 250 words, with the values in parentheses indicating the relative change compared to the previous setting.}\label{tbl9}
\end{table}

Table~\ref{tbl9} presents the classification performance of XLM-RoBERTa according to the number of extracted words. When using 50 words, both Micro-F1 and Macro-F1 scores were relatively low across all domains. When the number of words exceeded 200, a performance drop was observed in some domains.
Using 150 words proved to be the optimal setting, as it maintained strong performance in the hotel and beauty domains while minimizing the performance degradation in the food domain. Qualitative analysis also confirmed that the 150-word configuration produced the most balanced labeling results and helped prevent performance degradation caused by the inclusion of unnecessary terms. Based on these findings, we selected 150 words as the final optimal number for each aspect category.

\begin{center}
\begin{minipage}{0.95\textwidth}
\begin{tcolorbox}
[title={Aspect Category Candidates (Food Domain)},
colback = cBlue_1!10, colframe = cBlue_7, coltitle=white,
fonttitle=\bfseries\footnotesize, center title,
fontupper=\footnotesize, fontlower=\footnotesize]
Taste, Satisfaction, Service, Food Quantity, Menu, Waiting Time, Atmosphere, Cleanliness, Price, Purpose, Delivery, Location, Amenities
\end{tcolorbox}

\vspace{2mm}

\begin{tcolorbox}
[title={Aspect Category Candidates (Beauty Domain)},
colback = cBlue_1!10, colframe = cBlue_7, coltitle=white,
fonttitle=\bfseries\footnotesize, center title,
fontupper=\footnotesize, fontlower=\footnotesize]
Pigmentation, Persistence, Moisture, Skin Type, Spreadability, Usage Method, Scent, Color, Seasonal Use, Irritation, Ingredients, Hair, Improvement, Satisfaction, Packaging, Purchase
\end{tcolorbox}

\captionof{figure}{Food \& Beauty Aspect Category Candidates. The figure presents candidate aspect categories for the Food and Beauty domains, which are used for prompt-based assignment in our framework.}
\label{fig:food_beauty_aspect_category_prompt}
\end{minipage}
\end{center}

\setcounter{table}{0}
\section{Aspect Category \& Aspect Term}

This appendix presents the aspect categories selected in this study along with the corresponding lists of aspect terms. The aspect categories and associated aspect terms for the hotel, food, and beauty domains are provided in Table~\ref{tbl10}, Table~\ref{tbl11}, and Table~\ref{tbl12}, respectively.

\begin{table}[H]
\centering
\renewcommand{\arraystretch}{1.2} 
\footnotesize
\begin{tabular}{p{3.5cm} p{12cm}}
\noalign{\hrule height 1.3pt}
\toprule
\textbf{Aspect Category} & \textbf{Aspect Term} \\ 
\midrule
\textbf{룸 \footnotesize (Room)} & 세면대, 드라이기, 슬리퍼, 침실, 소파, 화장실, 어메니티, 커피포트, 칫솔, 조명, 면봉, 로션 \footnotesize \textit{(Washbasin, Hairdryer, Slippers, Bedroom, Sofa, Bathroom, Amenity, Coffee Pot, Toothbrush, Lighting, Cotton Swab, Lotion)} \\ 
\cmidrule(lr){1-2}
\textbf{목적 \footnotesize (Purpose)} & 여자친구, 남자친구, 엄마, 신혼여행, 휴가, 부모님, 우정, 여행, 호캉스, 연인, 커플, 기념, 신혼, 결혼기념일 \footnotesize \textit{(Girlfriend, Boyfriend, Mom, Honeymoon, Vacation, Parents, Friendship, Travel, Hotel Stay, Lover, Couple, Anniversary, Newlywed, Wedding Anniversary)} \\ 
\cmidrule(lr){1-2}
\textbf{위치 \footnotesize (Location)} & 공항버스, 재래시장, 이마트, 협재, 횟집, 해운대, 국제공항, 수산시장, 해수욕장, 미술관, 터미널, 관광지\footnotesize \textit{(Airport Bus, Traditional Market, Emart, Hyeopjae, Seafood Restaurant, Haeundae, International Airport, Fish Market, Beach, Art Museum, Terminal, Tourist Spot)} \\ 
\cmidrule(lr){1-2}
\textbf{부대시설 \footnotesize (Facilities)} & 레스토랑, 생맥주, 베이커리, 다이닝, 디너, 자판기, 뷔페, 주차장, 치킨, 무한리필, 한식, 스프, 미니바 \footnotesize \textit{(Restaurant, Draft Beer, Bakery, Dining, Dinner, Vending Machine, Buffet, Parking Lot, Chicken, Unlimited Refill, Korean Cuisine, Soup, Mini-bar)} \\ 
\cmidrule(lr){1-2}
\textbf{뷰 \footnotesize (View)} & 일몰, 풍경, 전경, 야경, 노을, 항구, 정원, 한라산, 해돋이, 감상, 바다, 전망, 뷰, 햇살, 오션\footnotesize \textit{(Sunset, Scenery, Panorama, Night View, Afterglow, Harbor, Garden, Hallasan, Sunrise, Appreciation, Sea, View, Sunshine, Ocean)} \\ 
\cmidrule(lr){1-2}
\textbf{서비스 \footnotesize (Service)} & 물어보다, 죄송하다, 말씀, 대처, 상담, 말씀드리다, 전화하다, 카운터, 요청, 프론트, 응대, 감사 \footnotesize \textit{(Inquire, Apologize, Communicate, Handle, Consult, Inform, Call, Counter, Request, Front Desk, Respond, Appreciate)} \\ 
\cmidrule(lr){1-2}
\textbf{만족도 \footnotesize (Satisfaction)} & 평범하다, 오래되다, 일반, 그럭저럭, 어수선하다, 노후하다, 협소하다, 양호하다, 퀄리티, 나쁘다, 열악하다, 뛰어나다, 럭셔리, 훌륭하다 \footnotesize \textit{(Average, Old, General, So-so, Disorganized, Outdated, Cramped, Good, Quality, Bad, Poor, Excellent, Luxurious, Superb)} \\ 
\bottomrule
\noalign{\hrule height 1.3pt}
\end{tabular}
\caption{Hotel Aspect Categories and Representative Terms (Korean with English Translation). 
These categories and representative terms are automatically derived by our framework from hotel review texts, and the Korean aspect terms are presented along with their corresponding English translations.}\label{tbl10}
\end{table}

\begin{table}[H]
\centering
\renewcommand{\arraystretch}{1.2} 
\footnotesize
\begin{tabular}{p{3.8cm} p{12cm}}
\noalign{\hrule height 1.3pt}
\toprule
\textbf{Aspect category} & \textbf{Aspect term} \\ 
\midrule
\textbf{음식 \footnotesize (Food)} & 포테이토, 베이컨, 머쉬룸, 발사믹, 고르곤졸라, 프로슈토, 에그베네딕트, 칼국수, 보쌈, 된장찌개 \footnotesize \textit{(Potato, Bacon, Mushroom, Balsamic, Gorgonzola, Prosciutto, Egg Benedict, Knife-cut Noodles, Bossam, Soybean Paste Stew)} \\  
\cmidrule(lr){1-2}

\textbf{음식량 \footnotesize(Food Quantity)} & 골고루, 여러가지, 세트, 글라스, 사이드, 무제한, 혼합, 플래터, 단일, 스몰, 메인, 글라스, 디시, 옵션 \footnotesize \textit{(Evenly, Various, Set, Glass, Side, Unlimited, Mixed, Platter, Single, Small, Main, Glass, Dish, Option)} \\  
\cmidrule(lr){1-2}

\textbf{맛 \footnotesize (Taste)} & 고소하다, 단맛, 조화롭다, 짭짤하다, 산뜻하다, 부드러움, 느끼하다, 달콤하다, 씁쓸하다 \footnotesize \textit{(Savory, Sweetness, Harmonious, Salty, Refreshing, Softness, Greasy, Sweet, Bitter)} \\  
\cmidrule(lr){1-2}

\textbf{만족도 \footnotesize (Satisfaction)} & 부실하다, 비싸다, 가격대비, 평범하다, 부족하다, 애매하다, 아쉬움, 실망하다, 저렴하다, 심각하다 \footnotesize \textit{(Inadequate, Expensive, Price-to-Value, Ordinary, Lacking, Vague, Regret, Disappointed, Affordable, Serious)} \\   
\cmidrule(lr){1-2}

\textbf{위치 \footnotesize (Location)} & 뚝섬역, 이대, 안국역, 홍대, 신촌, 성수동, 서울, 강남, 뚝도, 대학가, 을지로, 이태원 \footnotesize \textit{(Ttukseom Station, Ewha (Edae), Anguk Station, Hongdae, Sinchon, Seongsu-dong, Seoul, Gangnam, Ttukdo, University Area, Euljiro, Itaewon)} \\   
\cmidrule(lr){1-2}

\textbf{분위기 \footnotesize (Atmosphere)} & 아기자기하다, 화려하다, 예쁘다, 화사하다, 세련되다, 고급, 모던, 빈티지, 심플, 힙하다, 엔틱\footnotesize \textit{(Charming, Glamorous, Pretty, Bright, Sophisticated, Luxurious, Modern, Vintage, Simple, Hip, Antique)} \\
\cmidrule(lr){1-2}

\textbf{서비스 \footnotesize (Service)} & 말투, 표정, 아주머니, 서빙, 태도, 죄송하다, 종업원, 알바, 매니저, 응대, 불친절, 불쾌하다 \footnotesize \textit{(Tone of speech, Facial expression, Middle-aged woman, Serving, Attitude, Apologize, Staff, Part-timer, Manager, Customer service, Unfriendly, Unpleasant)} \\
\cmidrule(lr){1-2}

\textbf{목적 \footnotesize (Purpose)} & 남자친구, 남편, 와이프, 엄마, 아빠, 부모님, 동료, 결혼기념일, 신년, 가족, 생일, 외식, 명절 \footnotesize \textit{(Boyfriend, Husband, Wife, Mom, Dad, Parents, Colleague, Wedding Anniversary, New Year, Family, Birthday, Dining Out, Holiday)} \\
\cmidrule(lr){1-2}

\textbf{대기시간 \footnotesize (Waiting Time)} & 50분, 30분, 10분, 웨이팅, 기다림, 브레이크, 테이블링, 라스트, 오픈런, 평일 \footnotesize \textit{(50 minutes, 30 minutes, 10 minutes, Waiting, Waiting time, Break, Tabling, Last order, Open run, Weekday)} \\ 
\bottomrule
\noalign{\hrule height 1.3pt}
\end{tabular}
\caption{Food Aspect Categories and Representative Terms (Korean with English Translation). 
These categories and representative terms are automatically derived by our framework from food review texts, and the Korean aspect terms are presented alongside their corresponding English translations.}
\label{tbl11}
\end{table}

\begin{table}[H]
\centering
\renewcommand{\arraystretch}{1.2} 
\footnotesize
\begin{tabular}{p{3.3cm} p{12cm}}
\noalign{\hrule height 1.3pt}
\toprule
\textbf{Aspect category} & \textbf{Aspect term} \\ 
\midrule
Improvement & dehydrated, dryness, calm, aging, improved, plump, tigh, acneprone, smoothing, redness, youthful, hydrate, nourish, wrinkle, oiliness, glowing \\ 
\cmidrule(lr){1-2}
Ingredients & tocopheryl, ester, acetate, tocopherol, sativa, helianthus, palmitate, glycine, kernel, hydrogenated, panthenol, silica, annuus, argania, allantoin \\ 
\cmidrule(lr){1-2}
Satisfaction & actually, difficult, okay, hard, harder, good, well, great, nice, kind, tend, better, stronger, normal, properly, terrible, regular, exactly, strange, awesome, entirely \\ 
\cmidrule(lr){1-2}
Color & shimmer, highlighter, nude, beige, flittery, neon, coral, darker, bright, pink, gold, brown, bronze, rosy, color, yellow, blue, blond, gray, silver, smokey, navy, ashy \\ 
\cmidrule(lr){1-2}
Hair & straightener, wavy, tail, wave, curly, bun, bang, pony, curl, tangle, anchor, wig, shear, clip, tapered, turban, mane, dryer, heated, hair, shaft, trim, pin, snag \\ 
\cmidrule(lr){1-2}
Packaging & storage, sippered, tray, pouch, sleeve, soered, insert, zip, bag, tin, pocket, sealed, square, closure, foil, removable, encased, freebie, bowl, hook, screw, hole, ribbon \\ 
\cmidrule(lr){1-2}
Scent & pungent, fruity, flowery, musky, woodsy, spicy, floral, earthy, minty, sweet, citrusy, patchouli, sandalwood, vanilla, lavender, peppermint, tabacco, bergamot, jasmine, cedarwood \\ 
\cmidrule(lr){1-2}
Usage method & cleanse, pat, toner, massage, wipe, patted, rub, scrub, soak, showering, dab, rinsing, apply, wiped, exfoliate, spread, scrubbing, lather, absorbed, distribute, rinsed, lathering \\ 
\cmidrule(lr){1-2}
Purchase & bought, ordered, shopping, paid, acquired, picked, gifted, returned, exchanged, mom, sister, teen, son, dad, female, family, boyfriend, husband, kid, lady, boy, oldest, mail \\ 
\cmidrule(lr){1-2}
Persistence & lasted, extended, duration, progress, multiple, continuous, third, second, minimum, maximum, later, final, broke, fourth, third, repeated \\ 
\bottomrule
\noalign{\hrule height 1.3pt}
\end{tabular}
\caption{Beauty Aspect Categories and Representative Terms (Korean with English Translation). These categories and representative terms are automatically derived by our framework from beauty review texts, with the Korean aspect terms presented alongside their corresponding English translations.}\label{tbl12}
\end{table}

\setcounter{table}{0}
\section{Multi-Aspect Category Labeling Results}
\label{multi-aspect_category_labeling_data}

This appendix presents example results automatically labeled by the proposed \muscad~based on real-world review data.
For each domain (hotel, food, and beauty), ten review instances were selected, and the corresponding extracted multi-aspect categories are provided.
The results can be found in Tables~\ref{tbl13}, \ref{tbl14}, and \ref{tbl15}.

\begin{table}[H]
\centering
\renewcommand{\arraystretch}{1.6}
\setlength{\tabcolsep}{5pt} 
\footnotesize
\begin{tabular}{p{10.5cm} p{4.5cm}}
\noalign{\hrule height 1.3pt}
\toprule
\textbf{Review} & \textbf{Multi-Aspect Category} \\ 
\midrule
\parbox[t]{10.5cm}{
제주시내에 위치하고 있어서 편리하게 이용하기 좋아요\\
\textit{(It is conveniently located in Jeju City, making it easy to use.)}
} 
& 
\parbox[t]{4.5cm}{
만족도, 위치\\
\textit{(Satisfaction, Location)}
} \\ \hline

\parbox[t]{10.5cm}{
아침일출과 밤바다 보이는 바닷가 객실을 추천합니다\\
\textit{(I recommend the beachside room with views of the sunrise and the night sea.)}
} 
& 
\parbox[t]{4.5cm}{
뷰\\
\textit{(View)}
} \\ \hline

\parbox[t]{10.5cm}{
숙소 수영장 조식까지 최고였어요\\
\textit{(The accommodation, swimming pool, and breakfast were excellent.)}
} 
& 
\parbox[t]{4.5cm}{
만족도, 부대시설\\
\textit{(Satisfaction, Facilities)}
} \\ \hline

\parbox[t]{10.5cm}{
아이들을 위한 키즈카페와 다양한 부대시설 너무 맘에든 첫 숙소였어요\\
\textit{(This was our first hotel that we really liked because of its kids' cafe and various facilities for children.)}
} 
& 
\parbox[t]{4.5cm}{
목적, 부대시설\\
\textit{(Purpose, Facilities)}
} \\ \hline

\parbox[t]{10.5cm}{
여자친구와 500일 기념 및 생일 기념으로 호텔에 방문했습니다\\
\textit{(I visited the hotel to celebrate 500 days with my girlfriend and also for her birthday.)}
} 
& 
\parbox[t]{4.5cm}{
목적\\
\textit{(Purpose)}
} \\ \hline

\parbox[t]{10.5cm}{
넓기도 하고 욕조 또한 정말 좋습니다\\
\textit{(It is spacious and the bathtub is also excellent.)}
} 
& 
\parbox[t]{4.5cm}{
만족도, 룸\\
\textit{(Satisfaction, Room)}
} \\ \hline

\parbox[t]{10.5cm}{
위치는 올래시장과 이중섭거리 바로 옆이라 뭐 먹기는 최고임\\
\textit{(The location is excellent for dining as it is right next to Ollae Market and Lee Jung-seop Street.)}
} 
& 
\parbox[t]{4.5cm}{
만족도, 위치\\
\textit{(Satisfaction, Location)}
} \\ \hline

\parbox[t]{10.5cm}{
깨끗하고 친절한 직원들에 감동입니다\\
\textit{(I am impressed by the cleanliness and the friendly staff.)}
} 
& 
\parbox[t]{4.5cm}{
룸, 서비스\\
\textit{(Room, Service)}
} \\ \hline

\parbox[t]{10.5cm}{
피트니스센터 왜케 기구들 너무 좋나요\\
\textit{(The gym equipment at the fitness center is amazing.)}
} 
& 
\parbox[t]{4.5cm}{
만족도, 부대시설\\
\textit{(Satisfaction, Facilities)}
} \\ \hline

\parbox[t]{10.5cm}{
침대 더블 1개 싱글 2개 배역 어른 둘 초딩 하나 가족인데 각자 하나씩 편하게 잤어요\\
\textit{(We had one double bed and two single beds – perfect for our family of two adults and one child, as everyone slept comfortably.)}
} 
& 
\parbox[t]{4.5cm}{
만족도, 룸, 목적\\
\textit{(Satisfaction, Room, Purpose)}
} \\ 

\bottomrule
\noalign{\hrule height 1.3pt}
\end{tabular}
\caption{Examples of Hotel Reviews with Multi-Aspect Category Labels (Korean and English). 
Each review is presented in Korean alongside its English translation, and the corresponding multi-aspect category labels are automatically assigned by our framework. 
This table illustrates how a single review can be associated with multiple categories (e.g., Satisfaction, Location, Facilities), reflecting various aspects of the hotel experience.}
\label{tbl13}
\end{table}

\begin{table}[H]
\centering
\renewcommand{\arraystretch}{1.6}
\setlength{\tabcolsep}{5pt}
\footnotesize
\begin{tabular}{p{10.5cm} p{4.5cm}}
\noalign{\hrule height 1.3pt}
\textbf{Review} & \textbf{Multi-Aspect Category} \\
\noalign{\hrule height 0.8pt}

\parbox[t]{10.5cm}{제 입맛에는 딤섬 특히 맛났고 광동식 탕수육 좋았어요\\
\textit{(The dim sum was especially delicious to my taste, and the Cantonese-style sweet and sour pork was great.)}} 
& \parbox[t]{4.5cm}{음식, 맛\\\textit{(Food, Taste)}} \\
\hline

\parbox[t]{10.5cm}{프라이빗한 모임으로 제격이네요\\
\textit{(Perfect for a private gathering.)}} 
& \parbox[t]{4.5cm}{목적\\\textit{(Purpose)}} \\
\hline

\parbox[t]{10.5cm}{창덕궁역 뷰맛집 엄마랑 호캉스 여행기념으로 다녀왔어요\\
\textit{(A great restaurant with a view near Changdeokgung Station. I visited to celebrate a hotel staycation with my mom.)}} 
& \parbox[t]{4.5cm}{위치, 분위기, 목적\\\textit{(Location, Atmosphere, Purpose)}} \\
\hline

\parbox[t]{10.5cm}{연말 단체 모임 갔었는데 좋았어요. 양은 적음\\
\textit{(I went there for a year-end group gathering, and it was nice. However, the portions were small.)}} 
& \parbox[t]{4.5cm}{만족도, 목적, 음식량\\\textit{(Satisfaction, Purpose, Food Quantity)}} \\
\hline

\parbox[t]{10.5cm}{애견동반되서 강아지와 행복한 시간 보냈네요\\
\textit{(I was able to bring my dog and had a great time together.)}} 
& \parbox[t]{4.5cm}{목적\\\textit{(Purpose)}} \\
\hline

\parbox[t]{10.5cm}{밀크티 정말 맛있고 패션후르츠 음료도 아주 맛있습니다\\
\textit{(The milk tea was really delicious, and the passion fruit drink was also very tasty.)}} 
& \parbox[t]{4.5cm}{음식\\\textit{(Food)}} \\
\hline

\parbox[t]{10.5cm}{그런데도 예의 없이 손님에게 화내고 소리지르는 모습에 화가 나 리뷰를 작성합니다\\
\textit{(However, I was angered to see the staff rudely yelling at customers, which prompted me to write this review.)}} 
& \parbox[t]{4.5cm}{서비스\\\textit{(Service)}} \\
\hline

\parbox[t]{10.5cm}{웨이팅이 조금 있었지만 금방 빠졌어요\\
\textit{(There was a bit of a wait, but it cleared up quickly.)}} 
& \parbox[t]{4.5cm}{대기시간\\\textit{(Waiting Time)}} \\
\hline

\parbox[t]{10.5cm}{부모님도 좋아하시고 메뉴가 어른들 드시기에도 좋습니다\\
\textit{(My parents liked it, and the menu is well-suited for older guests.)}} 
& \parbox[t]{4.5cm}{만족도, 목적\\\textit{(Satisfaction, Purpose)}} \\
\hline

\parbox[t]{10.5cm}{커피랑 젤라또 전문이라고 해서 들어왔는데 전문답게 맛있어요\\
\textit{(I came in because they specialize in coffee and gelato, and it truly tasted professional.)}} 
& \parbox[t]{4.5cm}{음식\\\textit{(Food)}} \\
\noalign{\hrule height 1.3pt}
\end{tabular}
\caption{Examples of Food Reviews with Multi-Aspect Category Labels (Korean and English).
Each review is displayed in Korean along with its English translation, and the corresponding multi-aspect category labels are automatically assigned by our framework. 
This table illustrates how a single review can span multiple categories (e.g., Food, Taste, Location, Purpose), reflecting various facets of the dining experience.}
\label{tbl14}
\end{table}

\begin{table}[H]
\centering
\renewcommand{\arraystretch}{1.2}
\footnotesize
\begin{tabular}{p{11cm} p{4cm}}
\noalign{\hrule height 1.3pt}
\hline
\textbf{Review}                                                                                                                                                         & \textbf{Multi-aspect category}         \\ \hline
\raggedright These are extremely dull and will wreck your nails.                                                                                                        & Improvement                            \\ \hline
\raggedright And the dark brown is not brown, it's rust/copper.                                                                                                         & Color                                  \\ \hline
\raggedright Works wonders. They have other scents as well and I like all of them, but the classic is my favorite.                                                      & Scent, Satisfaction                     \\ \hline
\raggedright I found them on Amazon and bought a 3 pack. They are compact and easy to carry.                                                                            & Satisfaction, Purchase, Packaging       \\ \hline
\raggedright Wonderful bath salt - smells great.                                                                                                                        & Scent, Satisfaction, Usage Method       \\ \hline
\raggedright After I washed and dried my hair, I lifted parts and gave them a quick squirt, let them dry, combed them down a little and did a light spray over the rest of my hair. & Hair, Usage Method                     \\ \hline
\raggedright This complex is great and contains all you need to combat aging and fine lines.                                                                            & Ingredients, Satisfaction, Improvement  \\ \hline
\raggedright I did apply a small dab of shea butter to that spot and it no longer itches AT ALL.                                                                       & Satisfaction, Usage Method, Improvement \\ \hline
\raggedright The rose quartz is a very light pink and white... very pretty.                                                                                            & Usage Method, Color                     \\ \hline
\raggedright It's been about 5 weeks and my cracks are almost fully healed and even my pedicurist was shocked!                                                         & Persistence, Improvement               \\ \hline
\noalign{\hrule height 1.3pt}
\end{tabular}
\caption{Examples of Beauty Reviews with Multi-Aspect Category Labels (English).
Each review is presented in English, and the corresponding multi-aspect category labels are automatically assigned by our framework.
This table demonstrates how a single review can reflect various aspects (e.g., color, scent, satisfaction, improvement), offering insight into diverse beauty product experiences.}\label{tbl15}
\end{table}

\section{Details of Compared Unsupervised Aspect Extraction Methods}
\label{ae_method_details}
To evaluate the effectiveness of \textsc{MUSCAD}, we compare our model with nine representative unsupervised aspect and topic extraction methods, including probabilistic, hybrid, and embedding-based approaches. All baselines are implemented under the same number of aspect categories and aspect terms for fair comparison.

\begin{enumerate}[label=(\arabic*)]

\item \texttt{LDA} (Latent Dirichlet Allocation): the probabilistic topic model that assumes each document is represented as a mixture of latent topics, where each topic is characterized by a distribution over words. It serves as a fundamental baseline for unsupervised topic discovery~\citep{LDA}.

\item \texttt{BTM} (Biterm Topic Model): an extension of LDA that models global word co-occurrence patterns (biterms) across the entire corpus rather than within individual documents, improving performance on short texts~\citep{BTM}.

\item \texttt{NMF} (Non-negative Matrix Factorization): a deterministic topic extraction approach that factorizes the document-term matrix into two non-negative matrices representing topics and their associated word distributions. It provides interpretable results without probabilistic assumptions~\citep{NMF}.

\item \texttt{CTM} (Combined Topic Model): a hybrid topic model integrating contextual embeddings from pre-trained transformers into classical probabilistic topic modeling, enhancing both coherence and interpretability~\citep{CTM}.

\item \texttt{BERTopic} (Bidirectional Encoder Representations for Topic Modeling): a transformer-based neural topic model combining BERT embeddings with dimensionality reduction (UMAP), clustering (HDBSCAN), and class-based TF–IDF to generate semantically meaningful topics~\citep{BERTopic}.

\item \texttt{Top2Vec} (Topic2Vec Representation Model): jointly learns topic and document vectors in a shared semantic space using Doc2Vec embeddings, identifying topics as dense clusters of semantically similar documents~\citep{Top2Vec}.

\item \texttt{KeyBERT} (Keyword Extraction using BERT): a keyword extraction model leveraging BERT embeddings and cosine similarity between document and candidate word vectors to identify semantically relevant terms~\citep{KeyBERT}.

\item \texttt{SimCSE} (Simple Contrastive Sentence Embedding): a sentence embedding model trained via contrastive learning to produce semantically rich sentence representations, applicable to clustering and topic discovery~\citep{SimCSE}.

\item \texttt{ABAE} (Attention-Based Aspect Extraction): an attention-based neural aspect extraction model that learns unsupervised aspect-specific representations from sentences, serving as one of the earliest neural approaches for unsupervised aspect extraction~\citep{he-etal-2017-unsupervised}.                        

\end{enumerate}

\setcounter{figure}{0}
\section{Prompt}
\label{apd:prompt}
This appendix presents the prompts used for multi-aspect labeling with LLMs in the comparative experiment described in Section~\ref{llm_compare}. In this study, we designed appropriate prompts to evaluate whether LLMs can assign correct aspect categories to review sentences under zero-shot and few-shot settings. The prompts are shown in Fig.~\ref{fig:Zero-shot Hotel prompt}, Fig.~\ref{fig:Zero-shot Food prompt}, Fig.~\ref{fig:Zero-shot Beauty prompt}, Fig.~\ref{fig:Few-shot example}, and Fig.~\ref{fig:Beauty Few-shot example}.

\begin{tcolorbox}
    [title={Zero-shot Prompt (Hotel Domain)},
    colback = cBlue_1!10, colframe = cBlue_7,  coltitle=white,
    fonttitle=\bfseries\scriptsize,
    center title,fontupper=\scriptsize,fontlower=\scriptsize]
    
    \textbf{당신은 주어진 리뷰 문장을 읽고 아래의 aspect 중 하나 이상으로 분류하는 작업을 맡았습니다.} \\
    반드시 아래 aspect 목록에서만 선택하고, 제공되지 않은 키워드는 절대 포함하지 마세요. \\

    \vspace{1mm}
    \textbf{Aspect 목록:} 목적, 룸, 위치, 부대시설, 만족도, 뷰, 서비스
    \vspace{1mm}

    \textbf{Instruction:}
    \begin{itemize}[itemsep=2pt,topsep=2pt,leftmargin=*]
        \item 리뷰를 가장 잘 설명하는 하나 이상의 aspect를 선택하세요.
        \item 문맥과 세부 사항을 주의 깊게 읽고 적합한 측면을 판단하세요.
        \item 선택한 aspect 키워드만 쉼표로 구분하여 반환하세요. 설명이나 추가 텍스트는 포함하지 마세요.
    \end{itemize}

    \textbf{Labeling guideline:}
    \begin{itemize}[itemsep=2pt,topsep=2pt,leftmargin=*]
        \item \textbf{목적:} 숙소 방문 목적 (예: 신혼여행, 가족여행 등)과 관련된 표현  
        \item \textbf{룸:} 객실의 청결, 방음, 어메니티 등 객실 환경과 관련된 표현  
        \item \textbf{위치:} 숙소의 위치, 접근성, 주변 시설과의 거리 관련 표현  
        \item \textbf{부대시설:} 수영장, 조식 등 숙소 내 제공되는 부대시설과 관련된 표현  
        \item \textbf{만족도:} 숙박 전반에 대한 만족도 또는 가격 대비 가치 관련 표현  
        \item \textbf{뷰:} 창문, 발코니 등에서 보이는 전망과 관련된 표현  
        \item \textbf{서비스:} 직원의 친절도, 응대 태도, 서비스 품질 관련 표현  
    \end{itemize}

    \tcblower

    \textbf{You are tasked with categorizing the given review sentence into one or more of the aspects listed below.} \\
    You must select only from the provided aspect list and must not include any keywords that are not listed. \\

    \vspace{1mm}
    \textbf{Aspect List:} Purpose, Room, Location, Facilities, Satisfaction, View, Service
    \vspace{1mm}

    \textbf{Instruction:}
    \begin{itemize}[itemsep=2pt,topsep=2pt,leftmargin=*]
        \item Select one or more aspects that best describe the review.
        \item Carefully read the context and details to determine the most appropriate aspect(s).
        \item Return only the selected aspect keywords, separated by commas. Do not include explanations or additional text.
    \end{itemize}

    \textbf{Labeling guideline:}
    \begin{itemize}[itemsep=1pt,topsep=1pt,leftmargin=*]
        \item \textbf{Purpose:} Expressions related to the reason for visiting the accommodation (e.g., honeymoon, family trip)  
        \item \textbf{Room:} Expressions related to room conditions such as cleanliness, soundproofing, and amenities  
        \item \textbf{Location:} Expressions related to the accommodation’s location, accessibility, and proximity to nearby facilities  
        \item \textbf{Facilities:} Expressions related to in-house amenities such as swimming pools and breakfast  
        \item \textbf{Satisfaction:} Expressions indicating overall satisfaction with the stay or value for money  
        \item \textbf{View:} Expressions related to the scenery from windows, balconies, etc.  
        \item \textbf{Service:} Expressions related to staff friendliness, customer service, and service quality  
    \end{itemize}
\end{tcolorbox}
\captionof{figure}{Zero-shot Prompt for Hotel Aspect Category Labeling (Korean with English Translation).This figure shows a zero-shot prompt used by the LLM to categorize hotel-related review sentences into predefined aspects (e.g., purpose, room, location, facilities, satisfaction, view, service). The original instructions and labeling guidelines are in Korean, with an English translation provided for clarity.}
\label{fig:Zero-shot Hotel prompt}

\begin{tcolorbox}
    [title={Zero-shot Prompt (Food Domain)},
    colback = cBlue_1!10, colframe = cBlue_7,  coltitle=white,fonttitle=\bfseries\scriptsize,
    center title,fontupper=\scriptsize,fontlower=\scriptsize]
    
    \textbf{당신은 주어진 리뷰 문장을 읽고 아래의 aspect 중 하나 이상으로 분류하는 작업을 맡았습니다.} \\
    반드시 아래 aspect 목록에서만 선택하고, 제공되지 않은 키워드는 절대 포함하지 마세요. \\

    \vspace{2mm}
    \textbf{Aspect 목록:} 만족도, 맛, 목적, 분위기, 서비스, 위치, 음식, 음식량, 대기시간
    \vspace{2mm}

    \textbf{Instruction:}
    \begin{itemize}[itemsep=2pt, topsep=2pt, leftmargin=*]
        \item 리뷰를 가장 잘 설명하는 하나 이상의 aspect를 선택하세요.
        \item 문맥과 세부 사항을 주의 깊게 읽고 적합한 측면을 판단하세요.
        \item 선택한 aspect 키워드만 쉼표로 구분하여 반환하세요. 설명이나 추가 텍스트는 포함하지 마세요.
    \end{itemize}

    \textbf{Labeling guideline:}
    \begin{itemize}[itemsep=2pt, topsep=2pt, leftmargin=*]
        \item \textbf{목적:} 방문한 이유 또는 목적 관련 표현
        \item \textbf{분위기:} 인테리어, 조명, 뷰 등 분위기 관련 표현
        \item \textbf{위치:} 장소의 위치, 접근성, 교통 관련 표현
        \item \textbf{만족도:} 전반적인 경험에 대한 만족도 표현 포함, 가성비 포함
        \item \textbf{음식:} 음식의 종류, 품질, 음식명 관련 표현 포함
        \item \textbf{음식량:} 음식의 양, 주문 양과 관련된 표현
        \item \textbf{대기시간:} 대기시간, 줄 서는 시간 관련 표현
        \item \textbf{맛:} 음식의 구체적인 맛과 관련된 표현
        \item \textbf{서비스:} 직원의 태도, 응대 속도, 친절도 및 전반적인 서비스 표현
    \end{itemize}

    \tcblower

    \textbf{You are tasked with reading a given review sentence and categorizing it into one or more of the aspects below.} \\
    Make sure to select only from the provided aspect list and do not include any additional keywords. \\

    \vspace{2mm}
    \textbf{Aspect List:} Satisfaction, Taste, Purpose, Atmosphere, Service, Location, Food, Portion Size, Waiting Time
    \vspace{2mm}

    \textbf{Instruction:}
    \begin{itemize}[itemsep=2pt, topsep=2pt, leftmargin=*]
        \item Select one or more aspects that best describe the review.
        \item Carefully read the context and details to determine the most appropriate aspect(s).
        \item Return only the selected aspect keywords, separated by commas. Do not include explanations or additional text.
    \end{itemize}

    \textbf{Labeling guideline:}
    \begin{itemize}[itemsep=2pt, topsep=2pt, leftmargin=*]
        \item \textbf{Purpose:} Expressions related to the reason for visiting (e.g., special occasion, family gathering)  
        \item \textbf{Atmosphere:} Expressions related to interior design, lighting, or view  
        \item \textbf{Location:} Expressions about accessibility, transportation, and convenience of the place  
        \item \textbf{Satisfaction:} Expressions indicating overall experience satisfaction, including value for money  
        \item \textbf{Food:} Mentions of food type, quality, or specific dishes  
        \item \textbf{Food Quantity:} Expressions related to food quantity or portion sizes  
        \item \textbf{Waiting Time:} Mentions of wait times, queue lengths, or delays in service  
        \item \textbf{Taste:} Expressions describing the specific taste of food  
        \item \textbf{Service:} Expressions regarding staff attitude, responsiveness, friendliness, and overall service quality  
    \end{itemize}
\end{tcolorbox}

\captionof{figure}{Zero-shot Food Prompt (Korean with English Translation). This figure presents a zero-shot prompt used by the LLM to categorize food-related review sentences into predefined aspects (e.g., satisfaction, taste, purpose, atmosphere, service, location, food, portion size, waiting time). The instructions and labeling guidelines are originally written in Korean, with an English translation included for clarity.}
\label{fig:Zero-shot Food prompt}

\begin{tcolorbox}
    [title={Zero-shot Prompt (Beauty Domain)},
    colback = cBlue_1!10, colframe = cBlue_7,  coltitle=white,fonttitle=\bfseries\scriptsize,
    center title,fontupper=\scriptsize,fontlower=\scriptsize]
    
    \textbf{You are tasked with categorizing the given review sentence into one or more of the aspects listed below.} \\
    You must select only from the provided aspect list and must not include any keywords that are not listed. \\

    \vspace{2mm}
    \textbf{Aspect:} Color, Hair, Improvement, Ingredients, Packaging, Persistence, Purchase, Satisfaction, Scent, Usage Method
    \vspace{2mm}

    \textbf{Instructions:}
    \begin{itemize}[itemsep=2pt, topsep=2pt, leftmargin=*]
        \item Choose one or more aspects from the provided list that best describe the review.
        \item Do not include any aspects that are not in the provided list.
        \item Only return the aspect keywords separated by commas. Do not include any explanations or additional text.
    \end{itemize}

    \textbf{Labeling guideline:}
    \begin{itemize}[itemsep=2pt, topsep=2pt, leftmargin=*]
        \item \textbf{Satisfaction:} Expressions related to overall satisfaction with the experience.
        \item \textbf{Ingredients:} Mentions of product ingredients, raw materials, or composition.
        \item \textbf{Usage Method:} References to product usage methods or applications (e.g., face, body, cosmetics, etc.).
        \item \textbf{Improvement:} Aspects that have improved after using the product.
        \item \textbf{Color:} Opinions related to the product’s color.
        \item \textbf{Hair:} Effects on hair after using the product or changes related to hair condition.
        \item \textbf{Packaging:} Mentions of packaging design, durability, shipping condition, etc.
        \item \textbf{Scent:} Descriptions of fragrance, intensity, longevity, or types of scent.
        \item \textbf{Purchase:} Mentions of purchasing experience, reasons for purchase, or target audience for the product.
    \end{itemize}
\end{tcolorbox}

\captionof{figure}{Zero-shot Beauty Prompt (English Only).
This figure shows a zero-shot prompt used by the LLM to classify beauty-related review sentences into predefined aspects (e.g., color, hair, improvement, packaging, etc.).
All instructions, aspect lists, and labeling guidelines are provided in English, and users must select only from the given aspect categories without adding any unlisted keywords.}
\label{fig:Zero-shot Beauty prompt}

\begin{figure}[htbp]
\centering
{\scriptsize
\begin{tcolorbox}
    [title={Few-shot Example Prompt (Hotel Domain)},
    colback = cBlue_1!10, colframe = cBlue_7,  
    coltitle=white,fonttitle=\bfseries\scriptsize,
    center title, fontupper=\scriptsize]

    \begin{itemize}[itemsep=2pt, topsep=2pt, leftmargin=*]
        \item 리뷰: 깨끗하고 방도넓고 함덕해변도 가까워요  
        \quad → \quad \textbf{위치, 룸} \\
        The room was clean and spacious, plus Hamdeok Beach was just a short walk away!  
        \quad → \quad Location, Room

        \item 리뷰: 공항이랑 정말 가깝고 주변에 편의점이나 식당이 많아서 좋았습니다  
        \quad → \quad \textbf{위치, 만족도} \\ 
        Super close to the airport, and there were plenty of convenience stores and restaurants nearby, which was really convenient.  
        \quad → \quad Location, Satisfaction

        \item 리뷰: 주변에 이마트 스타벅스 맥도날드 등 편의시설 및 식당 많았구요  
        \quad → \quad \textbf{위치} \\
        Lots of places around like E-Mart, Starbucks, and McDonald's really easy to find anything you need!  
        \quad → \quad Location

        \item 리뷰: 프론트 직원분들도 정말 친절하시고 객실도 청결하고 좋았어요  
        \quad → \quad \textbf{룸, 서비스, 만족도} \\
        The front desk staff were super friendly, and the room was spotless and comfy!  
        \quad → \quad Room, Service, Satisfaction

        \item 리뷰: 화장실에서 락스냄새가 좀 심하긴했지만 방은 크고 넓고 좋아요  
        \quad → \quad \textbf{룸, 만족도} \\
        There was a strong bleach smell in the bathroom, but the room itself was big and comfortable.  
        \quad → \quad Room, Satisfaction

        \item 리뷰: 조식 또한 간단하면서 먹을만해서 만족스럽습니다  
        \quad → \quad \textbf{부대시설, 만족도} \\
        Breakfast was simple but decent, and I was happy with it.  
        \quad → \quad Facilities, Satisfaction

        \item 리뷰: 가족 커플 둘다 추천드려요  
        \quad → \quad \textbf{목적} \\
        Great for both families and couples—highly recommend!  
        \quad → \quad Purpose

        \item 리뷰: 주차장 들어가는 길 하드코어 한거 빼곤 괜찮았어요  
        \quad → \quad \textbf{부대시설} \\
        The road leading to the parking lot was a bit rough, but everything else was totally fine.  
        \quad → \quad Facilities

        \item 리뷰: 카운터 직원분이 너무 친절하고 시설도 다 좋았어요  
        \quad → \quad \textbf{만족도, 서비스} \\
        The staff at the counter were super nice, and the facilities were all in great condition.  
        \quad → \quad Satisfaction, Service
    \end{itemize}
\end{tcolorbox}

\vspace{1mm}

\begin{tcolorbox}
    [title={Few-shot Example Prompt (Food Domain)},
    colback = cBlue_1!10, colframe = cBlue_7,  
    coltitle=white,fonttitle=\bfseries\scriptsize,
    center title, fontupper=\scriptsize]
    
    \begin{itemize}[itemsep=2pt, topsep=2pt, leftmargin=*]
        \item 리뷰: 카페 인테리어도 이쁘고 2층은 뷰가 이뻤어요  
        \quad → \quad \textbf{분위기} \\
        The cafe interior was beautiful, and the view from the second floor was amazing.  
        \quad → \quad Atmosphere  

        \item 리뷰: 냄새도 안나고 고기도 부드럽고 이래저래 좋았습니다  
        \quad → \quad \textbf{맛, 만족도} \\
        No bad smell, the meat was tender, and everything was great overall.  
        \quad → \quad Taste, Satisfaction

        \item 리뷰: 점심시간 줄은 길고 웍이 하나라 요리 나오는 시간이 좀 걸려요  
        \quad → \quad \textbf{만족도, 대기시간} \\
        The lunch line is long, and since there's only one wok, food takes some time to come out.  
        \quad → \quad Satisfaction, Waiting Time

        \item 리뷰: 서울숲 올 일 있을때마다 들리고 싶네요  
        \quad → \quad \textbf{위치} \\
        I’d love to stop by whenever I visit Seoul Forest.  
        \quad → \quad Location  

        \item 리뷰: 버섯 전골 맛있어요  
        \quad → \quad \textbf{음식} \\
        The mushroom hot pot is delicious.  
        \quad → \quad Food  

        \item 리뷰: 생일메세지까지 섬세하게 챙겨주셔서 너무 감동받았어요  
        \quad → \quad \textbf{목적, 서비스} \\
        They even prepared a birthday message for me, which was so thoughtful and touching.  
        \quad → \quad Purpose, Service  

        \item 리뷰: 내부가 예쁘고 남자직원분 친절해요  
        \quad → \quad \textbf{분위기, 서비스} \\
        The interior is pretty, and the male staff member was very friendly.  
        \quad → \quad Atmosphere, Service  

        \item 리뷰: 크림 너무 부드럽고 담백 시트도 촉촉해요  
        \quad → \quad \textbf{맛} \\
        The cream is super smooth and light, and the cake layers are moist.  
        \quad → \quad Taste  

        \item 리뷰: 저랑 친구는 그래도 음식 3개를 시켜서 배부르게 먹고 왔습니다  
        \quad → \quad \textbf{만족도, 음식량} \\
        My friend and I ordered three dishes and left feeling full and satisfied.  
        \quad → \quad Satisfaction, Food Quantity  
    \end{itemize}
\end{tcolorbox}

\vspace{1mm}
} 

\caption{Few-Shot Examples for the Hotel \& Food Domains.
This figure demonstrates how examples of multi-aspect category labels assigned to reviews in the Hotel and Food domains are provided to the model in a few-shot setting.
By referring to the aspect categories assigned to each review, the model is guided to classify new sentences more accurately.}
\label{fig:Few-shot example}
\end{figure}

\begin{figure}[htbp]
\centering
{\scriptsize

\begin{tcolorbox}
    [title={Few-shot Example Prompt (Beauty Domain)},
    colback = cBlue_1!10, colframe = cBlue_7,  
    coltitle=white,fonttitle=\bfseries\scriptsize,
    center title, fontupper=\scriptsize]
    
    \begin{itemize}[itemsep=2pt, topsep=2pt, leftmargin=*]
        \item Review: Opened the package \& instant migraine. \quad → \quad \textbf{Improvement}  
        \item Review: Beautiful palette very pleased! \quad → \quad \textbf{Color}  
        \item Review: The price and deal can't be beat, lasts awhile. \quad → \quad \textbf{Satisfaction}  
        \item Review: My hair is so soft, I didn't need conditioner. \quad → \quad \textbf{Satisfaction, Improvement}  
        \item Review: Removes makeup and dirt like magic. \quad → \quad \textbf{Satisfaction, Usage Method}  
        \item Review: These socks are utterly amazing!!! \quad → \quad \textbf{Ingredients, Usage Method}  
        \item Review: Worked within a few hours. \quad → \quad \textbf{Persistence}  
        \item Review: Argan oil knows what they are doing! \quad → \quad \textbf{Satisfaction, Ingredients}  
    \end{itemize}
\end{tcolorbox}

} 
\caption{Few-Shot Examples for the Beauty Domains.
This figure demonstrates how examples of multi-aspect category labels assigned to reviews in the Beauty domains are provided to the model in a few-shot setting.
By referring to the aspect categories assigned to each review, the model is guided to classify new sentences more accurately.}
\label{fig:Beauty Few-shot example}
\end{figure}

\clearpage

\end{document}